\documentclass[11pt]{article}
\usepackage[letterpaper,margin=1in]{geometry}
\usepackage{times}
\usepackage[round,authoryear]{natbib}
\usepackage{amsmath,amssymb,amsthm,mathtools,booktabs,longtable,array,graphicx,xcolor,microtype}
\usepackage{flafter,xurl}
\usepackage[colorlinks=true,linkcolor=blue!55!black,citecolor=blue!55!black,urlcolor=blue!55!black]{hyperref}
\graphicspath{{figures/}}
\newtheorem{theorem}{Theorem}[section]

\newtheorem{proposition}[theorem]{Proposition}

\theoremstyle{remark}
\newcommand{\Fin}{\succeq_{\rm fin}}
\newcommand{\TV}{d_{\rm TV}}
\newcommand{\Hist}{\mathcal H}
\newcommand{\Q}{\mathcal Q}
\newcommand{\Def}{\delta}

\title{When Do Intrinsic Rewards\\Lead to Exploration?}
\author{%
  Scott W Viteri \quad Laura Gomezjurado Gonzalez \quad Clark Barrett\\[0.4em]
  Stanford University\\[0.3em]
  \small\texttt{scottviteri@gmail.com}\quad\texttt{lpgomez@stanford.edu}\\
  \small\texttt{barrettc@stanford.edu}%
}

\hypersetup{pdftitle={When Do Intrinsic Rewards Lead to Exploration?},
  pdfauthor={Scott W Viteri, Laura Gomezjurado Gonzalez, Clark Barrett}}
\date{}
\begin{document}
\maketitle

\begin{abstract}
Intrinsic rewards are designed to guide exploration in reinforcement learning by
assigning value to an agent's experience, for example through prediction error or
learning progress. However, maximizing these rewards need not produce the most
informative experience available. We propose a formal criterion for exploration that
compares policies by the counterfactual information they acquire: how well their
histories can substitute for experience under alternative policies. We construct a
single, simple environment in which specified count-based, prediction-error,
empowerment, and information-gain objectives have maximizing policies that are Pareto-suboptimal at
acquiring counterfactual information. We explain these failures and establish conditions
under which existing intrinsic rewards successfully encourage optimal exploration.
We also construct an objective that assigns a higher value whenever exploration strictly
improves under our criterion.
\end{abstract}

\section{Introduction}
Exploration concerns what an agent learns through interaction, including about
interactions it never performs. Intrinsic rewards can favor hard-to-predict observations
or changes in beliefs about unknown properties of the environment. We ask how the
evidence acquired by maximizing these rewards compares with what other ways of
acting could reveal.

\begin{figure}[!htb]
\centering
\includegraphics[width=\linewidth]{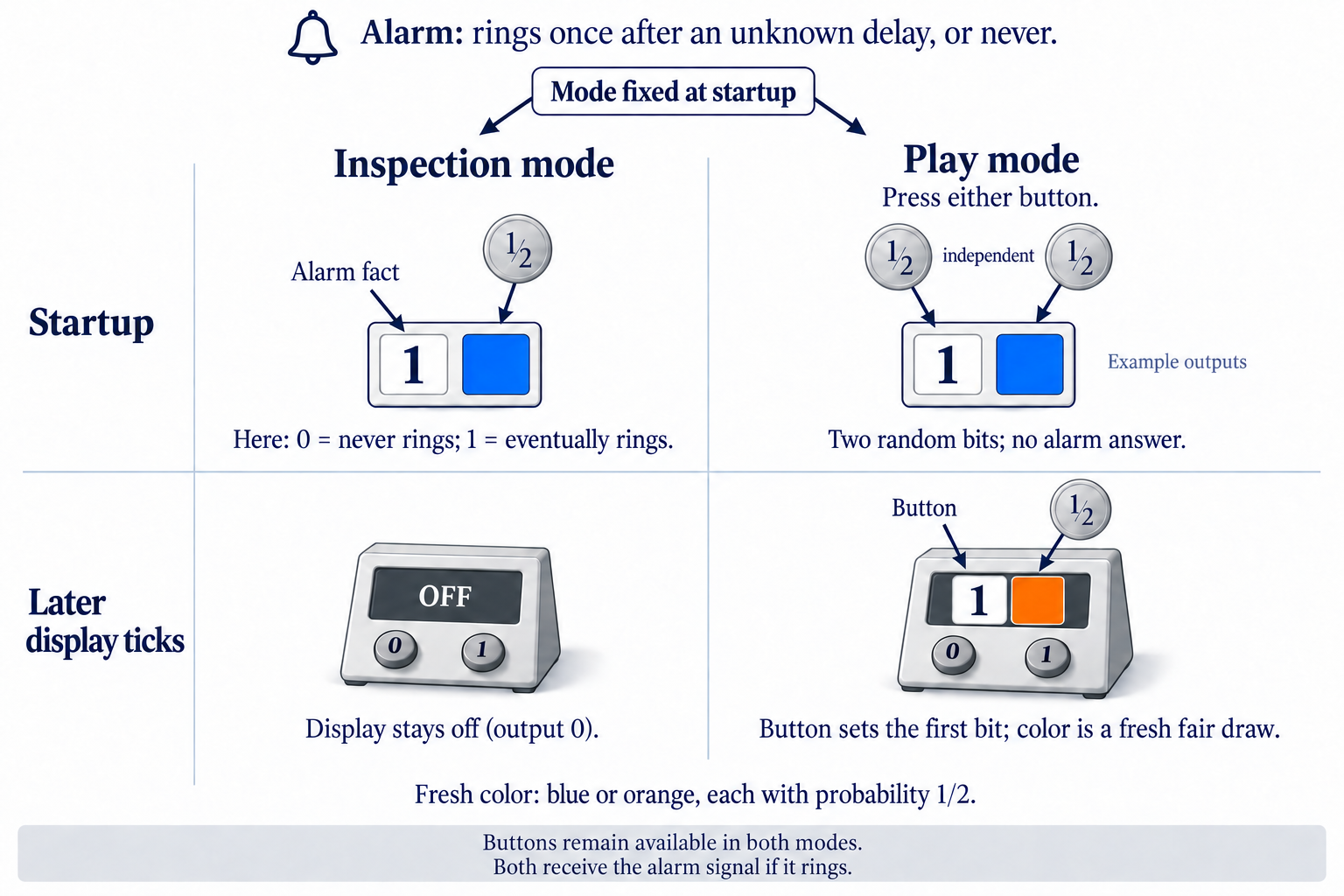}
\caption{\textbf{The alarm environment.}
Each digit--color pair encodes one observation; the displayed values are examples,
and coins denote independent fair draws. In inspection mode, the startup digit
reports whether the alarm will ever ring. The color distribution is known and identical
in every possible world. After startup, display observations alternate with alarm
observations, which report whether the alarm rings at that tick regardless of the
agent's actions.}
\label{fig:alarm-environment}
\end{figure}

Consider a machine with an alarm that rings once after an unknown delay or never rings.
A \emph{world} specifies one possible way the environment responds to actions.
Here the possible worlds differ only in whether and when the alarm rings, and
the agent does not know which is actual. At startup, the agent chooses between
two permanent modes: \emph{inspection} reveals whether the alarm will ever ring,
then leaves the display off; \emph{play} keeps it active without revealing that answer.
The buttons remain available and the alarm signal observable in both modes.
The active display responds to button presses with digits and colors according
to the same known probabilistic rule in every world.
Some intrinsic rewards favor rarely seen or hard-to-predict observations, so the
display's outputs can earn these rewards even though there is no unknown display
mechanism to learn. Figure~\ref{fig:alarm-environment} illustrates this startup choice.

We compare policies---rules for choosing actions from past actions and
observations---by the evidence they acquire. Running a policy for a fixed finite
number of steps produces a history of actions and observations. Each possible
world determines a distribution over that history; this family of distributions
is a \emph{statistical experiment}. We ask whether a randomized transformation
of one policy's collected history can reproduce the history distribution that
another policy would produce from the original starting point in the same world.
The transformation may depend on the alternative policy, its number of steps,
and the requested accuracy, but must achieve that accuracy in every possible
world without receiving the world label. Additional collection time is allowed.

An inspecting policy's history contains the startup answer and subsequent alarm
observations. A randomized transformation can follow an alternative policy's
action rule, generate display outputs from the known rule, and copy these alarm
observations. This reproduces the history distribution of a policy that keeps
the display active from startup, even though the inspecting agent's own display
remains off. Skipping inspection loses the ability to reproduce the startup
answer with small error uniformly across possible worlds: at every finite time,
silence is compatible with both a later alarm and no alarm.

Can a reward-maximizing policy be \emph{strictly dominated}---can another policy
reproduce all its finite-history distributions while the reverse fails?

We study the objective's optimal policies directly, separating their
incentives from the difficulty of finding them through learning. A reward need
not explicitly compare possible worlds to be evaluated this way: we evaluate
its expected return in those worlds while retaining its original inputs.
Our contributions are:
\begin{itemize}
\item A formal definition of exploration by the history distributions policies can
reproduce, used to characterize the exact optima of intrinsic objectives on the
alarm and separate examples.
\item Undiscounted cumulative information, Brier, and related belief-based objectives
reward every strict exploration improvement on finite classes with positive prior mass
at each world.
\item For finite action and observation sets and arbitrary world classes, a
representation of the comparison by countably many finite experiments and a bounded
score that rewards every strict improvement.
\item A finite planning study of how six objectives or controls trade off the
history distributions their selected policies can reproduce across 22 world classes.
\end{itemize}
We also give limits on the feasibility of complete exploration and the
computability of such scores.

Our comparison builds on statistical experiment comparison
\citep{blackwell1953equivalent,lecam1964sufficiency,torgersen1991comparison} and
established constructions of strictly increasing scalar scores
\citep{hack2022representing}.

\section{Comparing the evidence policies acquire}\label{sec:comparison}
The alarm distinguishes a varied observation stream from evidence about which
world is actual. We compare objectives by asking whether randomized transformations
of collected histories can reproduce the history distributions of alternative
interactions.

\subsection{From interaction histories to experiments}
Fix finite nonempty action and observation sets $A$ and $O$. A \emph{history}
of length $t$ is
\[
 h=((a_1,o_1),\ldots,(a_t,o_t))\in\Hist_t=(A\times O)^t.
\]
Write $h_r$ for its first $r$ pairs and $h(a,o)$ for one appended pair.
We represent a world $Q$ by its \emph{controlled-prefix behavior}: a collection of
nonnegative numbers $p_Q(h)$ giving the probabilities of observation sequences
when the actions in $h$ are imposed. These numbers satisfy
\[
 p_Q(\varnothing)=1,\qquad
 \sum_{o\in O}p_Q(h(a,o))=p_Q(h)\quad\text{for every $h$ and $a$}.
\]
Thus the possible next observations exhaust the probability at $h$ separately
for each proposed action; the actions themselves are choices, not random events
assigned probabilities by the world. At a history with $p_Q(h)>0$, the ratio
$p_Q(h(a,o))/p_Q(h)$ is the conditional probability of the next observation.
Responses specified after impossible histories do not affect this behavior.

Fix a nonempty class $\Q$ of worlds. A policy $\pi$ assigns action probabilities
$\pi(\cdot\mid h)$ to each history, using the same rule in every world; let $\Pi$
contain all such policies. Combining the policy's action probabilities with the
world's response probabilities gives its history distribution
$K_{\pi,t}(Q,h)=p_Q(h)\prod_{r=1}^t\pi(a_r\mid h_{r-1})$.
A \emph{statistical experiment} assigns a distribution on a finite output set to
each world, and its observed output is called a \emph{signal}. The experiment
$K_{\pi,t}$ therefore has the collected history $H_t$ as its signal. We call any
$K_{\sigma,n}$ with finite $n$ a \emph{finite native experiment}, because it is
available by acting through the environment's own interface. Every policy here
starts from the empty history. An alternative experiment begins at this same
starting point, rather than continuing the collected history.

\subsection{Reproducing another experiment}
A \emph{randomized decoder} $G$ transforms a source signal $x$ into an output $z$
with probabilities $G(z\mid x)$, following the same rule in every world. Applied
to an experiment $E$, it produces the distribution
$(GE)(Q,z)=\sum_x E(Q,x)G(z\mid x)$. Simulating a target experiment
means reproducing its signal distribution in each world.
For distributions $u,v$ on the same finite set, their \emph{total variation distance}
is $\TV(u,v)=\frac12\sum_z|u(z)-v(z)|$. We measure the error in reproducing a
target $T$ by choosing the best decoder and testing it in the worst world:
\begin{equation}\label{eq:deficiency}
 \delta(E,T)=\inf_G\sup_{Q\in\Q}\TV((GE)(Q),T(Q)).
\end{equation}
This quantity is the \emph{deficiency} from $E$ to $T$. Different targets may use
different decoders; reproducing their joint output requires treating that output
as one target. For any prior---a probability distribution on worlds---suppose
we use a signal to choose among finitely many options, with measurable payoffs
in $[0,1]$ depending on the choice and world. Replacing the target signal with
its decoded simulation reduces the best expected payoff by at most the uniform
simulation error
\citep{blackwell1953equivalent,lecam1964sufficiency,torgersen1991comparison}
(Appendix~\ref{app:structure}).

Longer histories project deterministically to shorter prefixes, so
$\delta(K_{\pi,t},T)$ cannot increase with $t$. We compare policies by asking
whether each fixed target prefix can be reproduced to arbitrary accuracy after enough collection:
\begin{equation}\label{eq:order}
 \pi\Fin\rho\quad\Longleftrightarrow\quad
 \forall n\ge0:\ \inf_{t\ge0}\delta(K_{\pi,t},K_{\rho,n})=0.
\end{equation}
The collection time and decoder may depend on the target length $n$ and the
requested accuracy, but must work uniformly over worlds. This reflexive,
transitive relation is the \emph{finitary process preorder}. We write
$\pi\succ_{\rm fin}\rho$ when $\pi\Fin\rho$ but $\rho\not\Fin\pi$, and say that
$\rho$ is \emph{strictly dominated}. Within a feasible policy family, a policy
is \emph{undominated} if no other policy in that family strictly dominates it.
A policy is \emph{natively sufficient}, or
\emph{completely exploring}, if it dominates every policy. Completeness concerns
reproducing the distributions of finite native experiments, without one common deadline
for all targets or requiring continued control of the environment. In particular,
it need not identify the world exactly on each run. Appendix~\ref{app:structure}
proves the preorder laws and the equivalent test using deterministic native targets.

\subsection{Selecting policies by exact objective optimization}\label{sec:selection}
An intrinsic \emph{objective} $J(\pi)$ assigns a score to a policy. We use
\emph{return} for an expected aggregate of rewards. For step rewards $r_u$, the \emph{finite return} is $\mathbb E\sum_{u=1}^H r_u$,
and the \emph{discounted return} is
$\mathbb E\sum_{u\ge1}\gamma^{u-1}r_u$ for a \emph{proper discount} $0<\gamma<1$.
For nonnegative rewards, the \emph{complete return} is
$\lim_{H\to\infty}\mathbb E\sum_{u=1}^H r_u$, possibly infinite.
The \emph{long-run average return} is
$\lim_{H\to\infty}H^{-1}\mathbb E\sum_{u=1}^H r_u$, when this limit exists.
Each aggregation defines its own objective,
while the exploration comparison always permits later collection beyond a
finite scoring horizon.

We study maximizers of a scalar objective over a stated policy family. Write
$J_Q(\pi)$ for its value in world $Q$. Where a world prior $\alpha$ is specified,
we use $J(\pi)=\mathbb E_{Q\sim\alpha}J_Q(\pi)$; in the three state-based comparisons
without a world prior, we use $J(\pi)=\inf_{Q\in\Q}J_Q(\pi)$.
Each policy uses the same action rule across worlds and receives no world label.
The native score in Section~\ref{sec:native} is defined directly from simulation
errors across the class.

We evaluate returns from the specified response laws while preserving
each reward's original inputs and assumptions about the environment and allowed
policies. In a Markov decision process (MDP),
the agent observes a state, and the next-state distribution depends only on
that state and its action. A world
specifies which transition law is actual, while state-based rewards score the
observed states. A \emph{posterior} gives the probabilities of possible
worlds after conditioning a prior on the collected history. Forecast and posterior
rewards retain that class and prior inside each $J_Q$. The comparison concerns
the evidence acquired by the selected policy; a learning algorithm's training
history may already contain additional evidence, so objective optimality is not
a claim about its training performance.

A score is \emph{weakly monotone} if $\pi\Fin\rho$ implies $J(\pi)\ge J(\rho)$,
and \emph{strictly monotone} if it also gives strict inequality whenever
$\pi\succ_{\rm fin}\rho$. Strict monotonicity excludes strictly dominated maximizers in
any feasible policy family; if a sufficient policy is feasible, exactly the
sufficient policies maximize the score. This gives a general selection guarantee,
while an objective without it may still select sufficient policies in particular
environments. We examine those successes alongside its failures.

\section{Objectives with dominated maximizers}\label{sec:alarm}
We compare intrinsic objectives by the evidence their maximizing policies acquire.
We begin with the alarm environment in Figure~\ref{fig:alarm-environment}. The unknown
world is $\theta\in\Theta=\mathbb N\cup\{\infty\}$, with
$\mathbb N=\{0,1,\ldots\}$. After startup, \emph{display ticks} alternate with
\emph{alarm ticks}. The alarm pulses at alarm tick $\theta$ when $\theta$
is finite and never pulses otherwise. Write $b=\mathbf1_{\{\theta<\infty\}}$,
where $\mathbf1$ is the indicator of the stated condition.
There are three actions, INSPECT, PLAY$_0$, PLAY$_1$, and four observation labels
$0,1,2,3$. The first action fixes the mode: INSPECT selects inspection mode;
either PLAY action selects play mode. At startup, INSPECT returns $2b+C$,
where $C$ is a fresh fair color bit; either PLAY action returns a uniform label,
independently of $\theta$. Subsequently, display observation $2k+2$ is zero in inspection
mode and $2\mathbf1_{\{a=\mathrm{PLAY}_1\}}+C$ in play mode, where $a$ is the
action at that step. Alarm observation $2k+3$
is $\mathbf1_{\{\theta=k\}}$ in both modes. All fresh bits are independent.
All three actions remain available in both modes; later actions cannot change
the mode or recover a skipped startup answer. The query is a stipulated sensor,
not a requirement to solve a halting problem. For rewards that depend on beliefs,
we assign probability one half to an alarm that never rings and one half to one
that eventually rings. Conditional on ringing, each successive delay has half
the probability of the preceding one. Thus the prior is
$\alpha(\infty)=\tfrac12$ and $\alpha(k)=2^{-(k+2)}$ for $k\ge0$.

Let $s$ be the policy's initial inspection probability and $F$ the one-observation inspection
experiment. Every randomized history policy, with arbitrary continuation, satisfies
\begin{equation}\label{eq:alarm-error}
 \delta(K_{\pi,t},F)=\frac{1-s}{2}\quad(t\ge1),\qquad
 D(\pi):=\inf_{t\ge1}\delta(K_{\pi,t},F)=\frac{1-s}{2}.
\end{equation}
Fix a finite collection time and compare a world in which the alarm never rings
with one in which it rings later than that time. If inspection was skipped,
the collected histories have the same distribution in these two worlds, although
the correct startup answers differ. Any decoder must then use the same answer
distribution for both, giving error at least one half in one of the worlds.
Inspection supplies the answer exactly on the branch of probability $s$, while
the skipped branch has probability $1-s$, giving the lower bound in
\eqref{eq:alarm-error}. Copying the answer on inspected histories and guessing
fairly on the others, with a fresh independent color, attains the bound.

Inspection also preserves everything needed for any other finite native
experiment: its startup answer and subsequent alarm observations determine the
world-dependent responses, and the decoder can generate display outputs from the
known rule while following the target policy from startup. It is therefore sufficient
and strictly dominates every policy with $s<1$. The late-alarm world in the lower
bound can change with collection time: each fixed finite alarm eventually becomes
visible, but no finite period of silence resolves the startup question uniformly
over all worlds. Appendix~\ref{app:alarm} gives the full decoder argument.

The comparisons below distinguish rewards for variation that reveals no unknown
world property, rewards for the path of belief changes, and finite-time score
advantages that disappear in the complete return. Count, predictive, and
empowerment rewards use this alarm; state-based objectives use separate MDPs.
Posterior rewards are compared on the alarm and on a separate four-bit example.

\subsection{Variation without evidence about the world}
We remain in the alarm environment, whose observations have four possible labels,
$0,1,2,3$. We first study a count-based exploration bonus from
\citet{bellemare2016unifying}. Their construction infers an effective occurrence
count, called a \emph{pseudo-count}, from how a predictor's probability for an
observation changes after seeing it again. The bonus is the inverse square root
of this count, so observations with smaller counts earn larger rewards.

For our example, let $N_h(o)$ count occurrences of label $o$ in the collected
history $h$, with $N_{\varnothing}(o)=0$. Our predictor adds one to each of the
four counts and normalizes: after $t$ observations it assigns probability
$(N_h(o)+1)/(t+4)$ to $o$, initially $1/4$ for every label. Its pseudo-count is
exactly $N_h(o)+1$, so the reward for next observing
$o$ is $1/\sqrt{N_h(o)+1}$. Appendix~\ref{app:alarm} gives the derivation.

Inspection leaves at most two nonzero observations in the entire history, whereas
keeping the display active sustains visits to more labels and therefore earns a larger
bonus over a sufficiently long scoring horizon. For every $H\ge64$, a maximum
is attained and every maximizing policy skips inspection; the appendix proves
this comparison uniformly over all later action choices for the specified
count model. Every policy's complete return is $+\infty$, and every policy's
long-run average return is zero. Both objectives therefore tie all policies.

Still on the alarm, we next examine rewards for observations that a predictor
finds surprising. Let $q(o\mid h,a)$ be the Bayesian forecast from the stated
prior and collected history $h$, after choosing action $a$. The per-step
\emph{Shannon} and \emph{Square} rewards of
\citet[Definitions~16--17]{aslanides2016thesis} are respectively
$-\log_2 q(o\mid h,a)$ and $-q(o\mid h,a)$.
Shannon pays for an observation's surprise; its expected reward under the
forecast is the forecast's \emph{entropy}. For a discrete distribution $p$,
this is $\operatorname H(p)=-\sum_xp(x)\log_2p(x)$, with $0\log_2 0=0$.
Square's expected reward under the forecast is $-\sum_o q(o\mid h,a)^2$,
so it also favors probability spread over several outcomes. Both select exactly
the policies that skip inspection, for every $0<\gamma<1$.
Their preference therefore persists as the discount approaches one: keeping the
display active earns prediction reward for its known randomness even without
learning about the world.
\citet[Section~5]{orseau2013universal} already observe attraction to noise for
Shannon and whole-history Square; here we characterize every optimum and its native
deficiency on the fixed alarm for every proper discount.
Appendix~\ref{app:predictive} proves this for the actual history-conditioned
forecasts and distinguishes predictive Square from posterior Brier gain.

\emph{Empowerment} measures how much influence the agent's actions can have on a
future observation. To quantify this influence, \emph{mutual information}
$I(X;Y)=\operatorname H(X)-\operatorname H(X\mid Y)$ measures the expected
uncertainty reduction about $X$ after observing $Y$; conditional entropy averages
over $Y$. On the alarm, fix an action-sequence length $n\ge1$. At a visited
history in a fixed world, a hypothetical \emph{probe} chooses $n$ actions in
advance and retains only the final raw observation. Empowerment is the maximum
mutual information between this sequence and that observation, over probability
distributions on the sequences \citep{klyubin2005empowerment}. Intermediate
observations do not guide the probe's actions. This probe defines the reward
at the visited history; the exploration comparison still uses experiments from startup.

After startup on the alarm, successive starting times alternate whether the
probe's final observation comes from the display or the alarm. Empowerment is
one bit at an active display tick and zero at an alarm tick or when the display is off.
Using it as the reward at each visited history gives a divergent total when
$s<1$, so we use a long-run average return: $(1-s)/2$ bits per interaction.
Exactly the policies with $s=0$ maximize this long-run average return and every proper discounted
return, for each fixed $n$. This rewards retaining the ability to change the
displayed digit. A decoder can reproduce the display's output distributions
without making the inspecting agent's display responsive again.
Appendix~\ref{app:alarm} gives the channel, return, and proofs.

The next three objectives score physical MDP states, so we evaluate them in
separate finite MDPs. Each observation is the current state. At startup in each
example, one action reveals an unknown bit in the transition law; alternatives
give observations independent of that bit.
The \emph{maximum occupancy principle} (MOP) rewards the entropy of the action
distribution at each state and of the next-state distribution given that state
and action \citep{ramirezruiz2024occupancy}. With unit weights in our five-state
example, we optimize over \emph{stationary} policies, whose action rule depends
only on the current state. Every optimum reads the unknown bit determining the
transition law with probability $1/3$.

\emph{State-visitation entropy} scores the entropy of a mixture of state
distributions across times $t\ge0$, weighted by $(1-\gamma)\gamma^t$
\citep{hazan2019provably}. In our separate three-state example, every optimum
chooses an uninformative random state and remains there.

\emph{Diversity Is All You Need} (DIAYN) samples a label, or \emph{skill}, once
per run; its action probabilities depend only on the current state and that label.
It rewards information about the skill in the state and the entropy of the action
distribution given the state and skill \citep{eysenbach2019diayn}.
In a separate five-state example, every optimal family of these policies reads
the unknown bit with probability at most $1/2$. We use the limiting state
probabilities across runs and the exact information objective; the result does
not assert convergence of a learned skill predictor.
These three comparisons maximize the worst-world objective from
Section~\ref{sec:selection}. Appendices~\ref{app:mop-mdp}--\ref{app:diayn}
give the MDPs and prove the stated optimizer deficiencies.

\subsection{The path of learning can outweigh the evidence acquired}
We next ask how splitting evidence across several belief updates affects the
total reward. We compare four rules that reward individual posterior updates,
first on the alarm and then in an example offering complete information at once
or less information in stages. Returning to the alarm, let $\beta_t$ be the
posterior on worlds after $H_t$, with $\beta_0=\alpha$. Write $q=\beta_{t-1}$
and $p=\beta_t$. The four rewards for one update are:
\[
\begin{array}{ll}
 r_{\rm info}(q,p)=\sum_Q p(Q)\log_2\!\frac{p(Q)}{q(Q)}, &
 r_{\rm Brier}(q,p)=\sum_Q(p(Q)-q(Q))^2,\\[2pt]
 r_{\rm Hel}(q,p)=\sum_Q(\sqrt{p(Q)}-\sqrt{q(Q)})^2, &
 r_{\rm Abs}(q,p)=\sum_Q|p(Q)-q(Q)|.
\end{array}
\]
In the information reward, a term with $p(Q)=0$ contributes zero.
On positive-probability histories, Bayesian updating gives $p(Q)=0$ whenever
$q(Q)=0$. Expected information gain
is expected Shannon-entropy reduction
\citep{lindley1956measure}. Brier gain comes from the quadratic forecast loss
\citep{brier1950verification}; Hellinger and Absolute use the posterior-movement
rewards of \citet[Section~3]{orseau2013universal}.

All four favor inspection on the alarm under every proper discount. A separate
sixteen-world example reveals their difference: FULL reveals four independent
fair bits at once, whereas PARTIAL reveals three sequentially and never reveals
the fourth. Hellinger and Absolute prefer PARTIAL for the complete return and every
$9/10\le\gamma<1$. Every optimum has eventual FULL deficiency $1/2$.
Information and Brier instead prefer FULL. The sequence of smaller updates can
earn more movement reward despite revealing strictly less. Appendix~\ref{app:posterior-movement}
gives the exact returns and the favorable alarm cases.

\subsection{A finite-time advantage can disappear in the complete return}
On the alarm, information and Brier favor inspection at every positive finite
scoring horizon and every proper discount. Nevertheless, every policy's complete
information return is two bits and its complete Brier return is $2/3$, so all policies
maximize these limiting objectives, including those with $D(\pi)=1/2$. Watching eventually detects each finite alarm, and after long silence the
remaining prior probability of a future alarm tends to zero. Average uncertainty
therefore vanishes even though some possible alarm still lies beyond every finite
collection time. Both complete returns are finite.

Optimizing discounted returns therefore differs from optimizing their limit:
information and Brier optimizers inspect as $\gamma\uparrow1$, while optimizing
the limiting score admits all policies. Hellinger and Absolute instead prefer
PARTIAL both near one and at one in the four-bit example. Both comparisons
matter: failure beyond an objective's effective horizon does not explain which
experiments its optima can reproduce under the stated return.
An entire infinite pulse stream determines whether a pulse ever occurs, while
no finite silent prefix resolves that question uniformly over worlds.
Proposition~\ref{prop:next-terminal-obstruction} gives a simplified alarm example
in which two policies have equivalent infinite-history experiments but strictly
ordered finite-history processes.

\begin{table}[t]
\centering
\caption{\textbf{How well do objective optima reproduce native experiments?}
Entries give all eventual target deficiencies $D$ attained by global maximizers
within the stated policy families. Zero means sufficiency here; positive values
mean strict domination by a sufficient policy. All later collection is allowed. MOP, state-visitation entropy, and
DIAYN use worst-world values; other rewards use their stated priors. Errors from
different examples are not a performance ranking. The eventual reference $J_w$ (Section~\ref{sec:native}) is undiscounted and
has $D=0$ in every example.}
\label{tab:reward-map}
\setlength{\tabcolsep}{4pt}
\begin{tabular}{@{}lcccc@{}}
\toprule
& \multicolumn{2}{c}{Alarm: inspection target}
& \multicolumn{2}{c}{Four bits: FULL target}\\
Objective & Complete return & $0<\gamma<1$ & Complete return & $9/10\le\gamma<1$\\
\midrule
Information & $[0,1/2]$ & $0$ & $0$ & $0$\\
Brier & $[0,1/2]$ & $0$ & $0$ & $0$\\
Hellinger & $0$ & $0$ & $1/2$ & $1/2$\\
Absolute & $0$ & $0$ & $1/2$ & $1/2$\\
\midrule
$J_w$ (eventual reference) & \multicolumn{2}{c}{$0$} & \multicolumn{2}{c}{$0$}\\
\midrule
\multicolumn{2}{@{}l}{Objective and example}
& \multicolumn{2}{l}{Return / conditions} & $D$\\
\midrule
\multicolumn{2}{@{}l}{Categorical pseudo-count; alarm}
& \multicolumn{2}{l}{Finite $H\ge64$} & $1/2$\\
\multicolumn{2}{@{}l}{Categorical pseudo-count; alarm}
& \multicolumn{2}{l}{Complete; long-run average} & $[0,1/2]$\\
\multicolumn{2}{@{}l}{Predictive Shannon; alarm}
& \multicolumn{2}{l}{Discounted; $0<\gamma<1$} & $1/2$\\
\multicolumn{2}{@{}l}{Predictive Square; alarm}
& \multicolumn{2}{l}{Discounted; $0<\gamma<1$} & $1/2$\\
\multicolumn{2}{@{}l}{Sensor empowerment; alarm}
& \multicolumn{2}{l}{\shortstack[l]{Long-run average or discounted;\\fixed $n\ge1$}} & $1/2$\\
\multicolumn{2}{@{}l}{MOP; five-state MDP, READ}
& \multicolumn{2}{l}{Discounted; unit weights} & $1/3$\\
\multicolumn{2}{@{}l}{\shortstack[l]{State-visitation entropy;\\three-state MDP, TEST}}
& \multicolumn{2}{l}{Discounted occupancy} & $1/2$\\
\multicolumn{2}{@{}l}{DIAYN; five-state MDP, READ}
& \multicolumn{2}{l}{Stationary; positive entropy weight} & $[1/4,1/2]$\\
\bottomrule
\end{tabular}
\end{table}

Compression progress rewards a reduction in total description length when the
compressor updates, evaluating the old and updated compressors on the same
collected history. In a separate two-button example, Appendix~\ref{app:compression}
gives one deficient complete-return optimum when the description language,
initialization, and supplied background are fixed before selecting the possible-world
class. The comparison remains open when that class is supplied to the compressor:
its representation and use can change the score.

\section{Posterior objectives on finite and compact world classes}\label{sec:positive}
We now ask when cumulative information and Brier rewards preserve every strict
improvement in exploration. The alarm failure involved infinitely many worlds;
first restrict to a finite nonempty class $\Q$ and a prior $\alpha$ with
\emph{full support}: $\alpha(Q)>0$ for every world. Let $\Delta(\Q)$ be the set
of probability distributions on $\Q$. The expected undiscounted cumulative reward of each
method is the expected change in a single posterior score: negative entropy
$p\mapsto\sum_Qp(Q)\log_2p(Q)$ for information, and $p\mapsto\sum_Qp(Q)^2$
for Brier. These are classical posterior information values
\citep{degroot1962uncertainty,gneiting2007strictly}.

To generalize these examples, consider a function $\Psi:\Delta(\Q)\to\mathbb R$.
It is \emph{convex}
if $\Psi(\lambda p+(1-\lambda)q)\le\lambda\Psi(p)+(1-\lambda)\Psi(q)$ for
$0<\lambda<1$, and \emph{strictly convex} if the inequality is strict whenever
$p\ne q$. Define the limiting expected posterior-score gain:
\begin{equation}\label{eq:posterior}
 J_\Psi(\pi)=\lim_{t\to\infty}\mathbb E_{\alpha,\pi}\Psi(\beta_t)-\Psi(\alpha).
\end{equation}
For continuous convex $\Psi$, these expectations increase and are bounded, so the
limit exists.

\begin{theorem}[Finite-world posterior success]\label{thm:complete-posterior-objectives}
For a finite nonempty class of worlds, finite nonempty action and observation
sets, a full-support prior, and a continuous strictly convex function $\Psi$,
$J_\Psi$ is strictly monotone for the finitary process preorder. In any feasible family containing a natively
sufficient policy, its maximizers are exactly the sufficient policies.
\end{theorem}
The proof reconstructs an experiment with small prior-average error when
decoding causes little posterior-score loss. Every world has probability at least
$a=\min_{Q\in\Q}\alpha(Q)>0$, so a decoder with prior-average error at most
$\varepsilon$ has error at most $\varepsilon/a$ in every world. Consequently,
prior-average and uniform finitary comparison coincide here. The theorem does
not assume that any policy can identify the actual world with certainty. A finite
world class alone does not guarantee that this is possible.

On an infinite world class, vanishing prior-average error need not imply
vanishing uniform error. Define \emph{prior-average deficiency} by
$\delta_\alpha(E,T)=\inf_G\int\TV((GE)(Q),T(Q))\,d\alpha(Q)$, using one decoder
across all worlds.
Replacing $\delta$ by $\delta_\alpha$ in~\eqref{eq:order} defines the
\emph{prior-average finitary process preorder}. The alarm's vanishing average uncertainty
leaves a uniform error of $1/2$ without inspection. Positive prior mass at every
world is therefore insufficient on its own.
One positive infinite-class case requires the behavior class to be \emph{compact}
under convergence of every finite prefix mass: every sequence has a subsequence
whose masses converge to a behavior in the class. On such a class, a prior with
positive mass at every individual world makes the average and uniform finitary
preorders coincide. If the prior also has finite entropy, complete information is
finite and strictly monotone for the uniform preorder. These assumptions restrict the class to be countable.
Appendix~\ref{app:structure} proves this result and gives the precise
prior-average preorder characterization.

\section{A score that preserves every strict improvement}\label{sec:native}
The posterior guarantees require assumptions relating average and uniform errors.
For any nonempty world class, we can instead score the errors in reproducing
native experiments directly. To combine these errors into one score, we need a
list of targets that can detect every strict improvement in exploration. A
finite list can miss an improvement. Call a countable family $(T_j)_{j\ge1}$
of finite native experiments \emph{rich} if every
finite native experiment is approximated arbitrarily closely, on the same output
set and uniformly over worlds, by a member of the family. Such a family exists:
at each finite horizon only finitely many action distributions matter, and rational
approximations to their probabilities approximate the history law uniformly.
Write $\ell_j(\pi)=\inf_t\delta(K_{\pi,t},T_j)$ for the resulting eventual errors.

\begin{theorem}[Native representation and score]\label{thm:strict-finitary-objective}
For finite nonempty action and observation sets and any nonempty class of worlds,
a rich countable family of finite native experiments satisfies
\[
 \pi\Fin\rho\quad\Longleftrightarrow\quad
 \ell_j(\pi)\le\ell_j(\rho)\quad\text{for every }j.
\]
For any $w_j>0$ with $\sum_{j\ge1}w_j=1$, the \emph{native score}
\begin{equation}\label{eq:native-score}
 J_w(\pi)=1-\sum_{j\ge1}w_j\ell_j(\pi)
\end{equation}
is bounded in $[0,1]$ and strictly monotone for that order.
\end{theorem}
The weights set priorities among the simulation errors for different targets. The weighted-sum
construction is standard \citep{hack2022representing}; the native representation
supplies its coordinates. By Section~\ref{sec:selection}, attained maximizers are
undominated; when a sufficient policy is feasible, exactly the sufficient policies
maximize $J_w$, with value one. This gives a zero-deficiency reference in every
example in Table~\ref{tab:reward-map}.

Complete exploration is not always feasible. In a four-world example with two
unknown bits, the first action reveals one chosen bit and all later observations
are zero. Choosing the first bit with probability $s$ gives eventual errors
$(1-s)/2$ and $s/2$ for the respective reading experiments. Every policy is
undominated, yet none makes both errors zero (Appendix~\ref{app:feasibility}).
A strictly monotone score can select a compromise, but cannot remove this constraint.

Evaluating the score poses a separate problem. Appendix~\ref{app:computability}
gives two deterministic computable worlds in which no algorithm can evaluate
every computable policy's real score to arbitrary precision if the score is weakly
monotone and strictly prefers revealing the unknown bit at startup to always
waiting. Restricted settings may still permit effective optimization.

\section{Comparing finite information-gathering choices}\label{sec:experiments}
We test whether objectives optimized for short interactions acquire evidence
useful for reproducing the history distributions of longer interactions on the
same world class. We compute three-step policies from supplied response laws for
22 classes of four or eight worlds. Policies use their action--observation histories
without a world label; objectives requiring a prior use the uniform distribution
over worlds. We compare information, Brier, categorical pseudo-count, and two native
objectives with an independent uniform-action baseline. \emph{Native mean} and
\emph{native maximum} minimize the mean and largest simulation errors across all
128 deterministic three-step targets. We evaluate each selected experiment against
the hardest of all 32,768 four-step targets, excluded from optimization. Even at a
fixed target depth, minimizing mean error would not guarantee minimizing maximum
error. These are finite planning objectives, not the eventual score $J_w$; no
sampled training is used. Each reward retains its stated inputs, including raw
observations for the count bonus.

Across saved representative policies, native mean has lower four-step evaluation
error than ordinary Brier in 18 of 22 classes, with two ties and two losses.
But another native objective often
improves on native mean: minimizing native maximum while retaining a fixed
native-mean policy's Brier return yields 15 wins, one numerical tie, and six
losses against that reference. This second comparison uses fixed references,
so the difference in counts does not
isolate the reward constraint.
Targeting simulation error often helps here, but its aggregation matters. Numerically, in 21 of the 22 reward-matched pairs, neither experiment
can simulate the other exactly.
Appendix~\ref{app:finite-comparison} gives the full comparisons and asks whether
reproducing a reference within a fixed error tolerance requires sacrificing
return or is possible at numerically negligible cost. The presentation and
controls were designed after inspecting these cases; the results establish neither universal superiority nor
eventual exploration.

\section{Related work and limits}
\citet{schmidhuber2010formal} distinguishes prediction error from improvement in
prediction or compression, while \citet{bellemare2016unifying} connect count
bonuses to changes in a probability model.
\citet[Theorem~7]{orseau2013universal} give prediction of alternative continuations from the
reached history, under its prior and horizon assumptions and existence of an
optimal information-gain policy. Our targets begin at the initial interface,
including opportunities already lost, and require uniform simulation over worlds
rather than convergence in each fixed world.

The simulation criterion uses Blackwell comparison and Le Cam deficiency
\citep{blackwell1953equivalent,lecam1964sufficiency,torgersen1991comparison}.
The representation analysis of \citet{vanrooyen2014lecam} uses prior-weighted
deficiency to preserve decision value and includes active-learning experiments.
Sequential experiment comparison \citep{greenshtein1996comparison} preserves
information at prescribed decision times, whereas our comparison allows
additional collection time for each fixed target. Experiment simulation and positive weighted scalarization are established tools.
Reward-free exploration gives acquisition and
subsequent planning guarantees in finite episodic MDPs \citep{jin2020rewardfree};
we study what an objective values when optimization succeeds.

\begin{samepage}
We characterize which intrinsic optima lose counterfactual information and construct
a score preserving every strict exploration improvement. We analyze exact optimization;
learning remains separate. The native score depends on the interface and weights and
has no general computable evaluator. The earlier sixteen-target benchmark failed its
prespecified screen: no wins over Brier (Appendix~\ref{app:finite-study-scope}); broad
empirical superiority is unestablished. Validating world-class assumptions linking observed and
unperformed interactions and developing tractable objectives remain open.
\par
\end{samepage}

\section*{Acknowledgements}
This work was supported in part by the Stanford Center for Automated Reasoning.
Scott W Viteri also acknowledges support through Manifund for earlier exploratory
research that helped shape the ideas in this paper.

\section*{Ethics statement}
This work consists of theoretical analysis and experiments in synthetic environments.
It does not involve human participants or personal data.
\section*{Reproducibility statement}
The appendices give the mathematical proofs, exact objective definitions, and
all assumptions used in the comparisons. Appendix~\ref{app:finite-comparison}
specifies the finite study's response laws, random seeds, planning objectives, evaluation
budgets, numerical tolerances, and adverse results. Its verification subsection
distinguishes numerical certificates from the exact rational-table proof;
Appendix~\ref{app:verification} explains the scope of the accompanying formal
proofs. The companion code and data artifact (\url{https://github.com/scottviteri/what-is-exploration/tree/arxiv-v1}) provides experiment code,
selected data and numerical certificates, Lean proof sources, and reproduction
instructions. The study evaluates selected policies under supplied response laws
without training from sampled experience.
\section*{AI usage statement}
This project grew out of several years of thinking by the lead author. Its recent
development was greatly accelerated through collaboration with AI assistants,
including GPT-6 Astra, GPT-5.6 Sol, Claude Fable 5 and 5.1, and Claude Opus 5.
AI contributed throughout the research process, including mathematical framing,
formulation of claims, proofs, experimental design, implementation of experiments
and generation of experimental data through execution of AI-written code,
interpretation of results, exposition, and generation of the schematic in
Figure~\ref{fig:alarm-environment}. The Lean formalization was AI-generated.
The authors take responsibility for the final manuscript and accompanying
artifacts, including any errors.
\clearpage
\bibliographystyle{plainnat}
\bibliography{intrinsic_rewards/refs,intrinsic_rewards/rewrite_references}
\clearpage
\appendix
\section{An irreversible choice with no sufficient policy}\label{app:feasibility}
There are four deterministic worlds $Q=(b_1,b_2)\in\{0,1\}^2$ and two actions.
The first action chooses and returns one bit. Every later observation is zero,
regardless of the action. Policies may randomize and use the entire history.
Let $T_i$ be the one-step experiment that reads bit $i$.

\begin{proposition}[Two competing native targets]\label{prop:two-bit-infeasibility}
If $s$ is a policy's probability of initially reading bit 1, then, for every
positive collection time $t$,
\[
 \delta(K_{\pi,t},T_1)=\frac{1-s}{2},\qquad
 \delta(K_{\pi,t},T_2)=\frac{s}{2}.
\]
Consequently no policy is natively sufficient.
\end{proposition}
\begin{proof}
Every positive-length history retains the initial action and answer. Conversely,
from that pair a decoder can simulate all later actions using the known policy
and append the constant observations. The experiments observing the initial pair
and the full finite history therefore simulate each other exactly.
To reproduce either target, return the observed bit on the branch that read it
and guess fairly on the other branch. These decoders give the displayed upper
bounds. For the first lower bound, compare two worlds that differ only in bit 1.
Their target laws have total variation one and their source laws have total
variation $s$. Contractivity under decoding and the triangle inequality give
$1\le s+2\varepsilon$ for any uniform decoding error $\varepsilon$.
Exchanging the bits proves the second lower bound. At least one of the two
limiting errors is at least $1/4$, so no policy simulates every native target.
\end{proof}
These errors also distinguish every two different initial reading probabilities.
If $s<s'$, the policy with probability $s'$ has smaller error for $T_1$ but larger
error for $T_2$, so neither policy dominates the other. Policies with the same
$s$ have identical initial experiments and can reproduce each other's finite-history
distributions using their known action rules. Consequently every policy is undominated,
even though none is sufficient. The example uses no prior.

\section{Structural proofs}
\label{sec:next-structural-proofs}
\label{app:structure}

The main comparison asks whether maximizing an intrinsic objective can select a
policy that misses a strict improvement in reusable evidence. This appendix gives the
representation and posterior arguments supporting that comparison. Throughout,
a decoder is a randomized map that does not receive the unknown world as an
input. We write $\delta(E,F)$ for the uniform deficiency from experiment $E$
to experiment $F$, and $K_{\pi,t}$ for the experiment whose signal is policy
$\pi$'s full history at time $t$.
If a decoder applied to $E$ reproduces $F$ exactly in every world, $F$ is
a \emph{garbling} of $E$, and $E$ \emph{Blackwell dominates} $F$. This
dominance is strict when the reverse exact simulation is impossible.
In the relation $\pi\succeq_{\mathrm{fin}}\rho$, both policies start from the
empty history in the same world. At every finite target length, randomized
decoders of sufficiently long $\pi$-histories must reproduce $\rho$'s history
distributions to arbitrary accuracy uniformly over worlds.
For an infinite class, we use the smallest sigma-algebra on $\Q$ making every
controlled-prefix probability $Q\mapsto p_Q(h)$ measurable. This specifies the
sets to which priors assign probabilities and makes each finite experiment
measurable. Decision payoffs are assumed measurable in the world, so their
prior expectations are defined.

The analysis of \citet{vanrooyen2014lecam} uses prior-weighted deficiency for representations
that preserve downstream decision value and also discusses active-learning
experiments. \citet{liang2022dynamically} obtain acquisition-strategy comparisons
under a supplied Gaussian prior, a scalar payoff-relevant quantity and
stopping-payoff assumptions. These are precedents for reusable information;
our comparison uses eventual uniform simulation of finite native experiments.

The comparison is reflexive because the identity decoder preserves every history
distribution. To prove transitivity, fix a final target prefix and an accuracy $\varepsilon>0$.
First choose a middle-policy prefix simulating that target within
$\varepsilon/2$, then a source-policy prefix simulating the chosen middle
prefix within $\varepsilon/2$. Decoder composition and the deficiency triangle
inequality give the required final error. Longer histories preserve this bound
by prefix projection.

For completeness, deterministic and randomized finite native experiments impose the same zero-error-limit requirement. A deterministic contingent plan specifies one action at every possible history.
At any fixed depth there are finitely many such plans. Every randomized
policy's history distribution at that depth is a world-independent mixture of
these plans' history distributions: draw all its finitely many action choices
in advance. This is the finite behavioral-to-mixed strategy construction of \citet[Lemma 3, Equation (6)]{kuhn1953extensive};
the product weights depend only on the policy, so the same mixture works in every world.
If each plan
is simulable to the requested accuracy, choose one source time large enough
for all of them, draw the mixture index independently, and apply its decoder.
The mixture error is no larger than the largest component error. The reverse
implication holds because deterministic plans are policies. Thus simulating every deterministic finite native experiment to arbitrary
accuracy is equivalent to native sufficiency.

\subsection{Why countably many native targets suffice}

For a fixed target experiment $T$, define its eventual error under $\pi$ by
\[
 \ell_T(\pi)=\inf_t\delta(K_{\pi,t},T).
\]
Longer histories can reproduce shorter histories by deleting their suffix, so
these deficiencies decrease with $t$ and their infimum is their limit.
To compare arbitrary randomized policies, the target family must include
approximations to randomized native experiments, with their actions retained.

\begin{theorem}[Native representation and a strict score]
\label{thm:next-native-representation}
For any nonempty class of controlled-prefix behaviors with finite nonempty
action and observation alphabets, there is a countable sequence $(T_j)$ of
actual finite native experiments such that
\[
 \pi\succeq_{\mathrm{fin}}\rho
 \quad\Longleftrightarrow\quad
 \ell_{T_j}(\pi)\le\ell_{T_j}(\rho)\quad\text{for every }j.
\]
For any strictly positive weights $w_j$ summing to one, the native score
\[
 J_w(\pi)=1-\sum_j w_j\ell_{T_j}(\pi)
\]
lies in $[0,1]$, preserves the finitary process preorder, and strictly increases at every
strict finitary improvement. Every attained maximizer is undominated in its
feasible policy family. If that family contains a natively sufficient policy,
its maximizers are exactly its natively sufficient members.
\end{theorem}

\begin{proof}
Fix a history length $n$. A policy's behavior through time $n$ is determined by
finitely many action distributions, one for each history shorter than $n$.
This finite product of probability simplices has a countable dense subset.
The map from policies to history distributions is continuous uniformly over
the world class. Indeed,
if corresponding action distributions differ by at most $\eta$ in total
variation, draw each pair of actions jointly so that they disagree with
probability at most $\eta$ while the histories agree. After equal actions, use
the same world response. The probability of any action disagreement
by time $n$ is at most $n\eta$. Thus the resulting history laws differ by at
most $n\eta$ in every world. Enumerating dense policy tables at every depth
supplies the required countable family of native experiments.

Suppose first that $\pi\succeq_{\mathrm{fin}}\rho$. The deficiency triangle
inequality gives
\[
 \delta(K_{\pi,t},T_j)
 \le \delta(K_{\pi,t},K_{\rho,n})+\delta(K_{\rho,n},T_j).
\]
Take the source-time limit, then the infimum over $n$, to obtain
$\ell_{T_j}(\pi)\le\ell_{T_j}(\rho)$.

Conversely, suppose these coordinate inequalities hold. Fix $n$ and a positive
number $\eta$. Choose a target $T_j$ whose rows are within $\eta$ of
$K_{\rho,n}$, uniformly over worlds. Since $\rho$ contains this prefix,
$\ell_{T_j}(\rho)\le\eta$. Target perturbation and the assumed inequality give
\[
 \inf_t\delta(K_{\pi,t},K_{\rho,n})
 \le\ell_{T_j}(\pi)+\eta
 \le\ell_{T_j}(\rho)+\eta\le2\eta.
\]
Letting $\eta$ decrease to zero proves the reverse implication.

All losses lie in $[0,1]$. A strict comparison improves at least one coordinate
strictly and all others weakly, so its positive-weight sum improves strictly.
A maximizing policy therefore cannot have a strict dominator in its feasible
family. A sufficient feasible member dominates every policy. Strictness then
forces any maximizer to dominate it in return; this is exactly native
sufficiency. Monotonicity gives the converse.
\end{proof}

For each scored target, nonnegativity of the other losses gives
$w_j\ell_{T_j}(\pi)\le1-J_w(\pi)$, and hence
\[
 \ell_{T_j}(\pi)\le\frac{1-J_w(\pi)}{w_j}.
\]
This controls eventual error along scores approaching one, but supplies no
common acquisition deadline. If no sufficient policy is feasible, a maximizing
policy can have a nonzero profile and remain incomplete.

\subsection{Complete information and average simulation}

Fix a probability prior $\mu$ on the measurable world class. To distinguish
average from uniform simulation, define
\[
 \delta_\mu(E,F)=\inf_G\int
 \operatorname{TV}\bigl(GE(\theta),F(\theta)\bigr)\,d\mu(\theta).
\]
The same decoder $G$ appears inside every world's error. Define
$\pi\succeq_\mu\rho$ by requiring
$\inf_t\delta_\mu(K_{\pi,t},K_{\rho,n})=0$ for every $n$.
Uniform comparison implies this prior-average comparison because an average
error cannot exceed the largest error.

Write $\Theta$ for the random world drawn from $\mu$ in this subsection.
Let $I_\mu(\Theta;K_{\pi,t})$ denote mutual information, measured in nats,
between the world drawn from $\mu$ and the policy history. Define complete
information by
\[
 J_I(\pi)=\sup_t I_\mu(\Theta;K_{\pi,t})\in[0,\infty].
\]
The supremum retains the value infinity when prefix information is unbounded.

\begin{theorem}[Information equality detects reverse average simulation]
\label{thm:next-information-equality}
Suppose finite history alphabets are nonempty and their world-indexed
probabilities are measurable. If $\pi\succeq_\mu\rho$ and $J_I(\pi)<\infty$,
then
\[
 J_I(\pi)=J_I(\rho)
 \quad\Longleftrightarrow\quad \rho\succeq_\mu\pi.
\]
More quantitatively, for every fixed $n$,
\[
 \inf_m\delta_\mu(K_{\rho,m},K_{\pi,n})
 \le \min\!\left\{1,
 \sqrt{\frac{J_I(\pi)-J_I(\rho)}{2}}\right\}.
\]
No prior-support or identifying-policy hypothesis is required.
\end{theorem}

\begin{proof}
For an exact garbling $E\to F$, a Bayes reverse decoder samples the source
signal from its conditional law given the garbled signal under the prior.
The information loss is the conditional information discarded by this
operation. For finite probability vectors, relative entropy is
$D(p\Vert q)=\sum_xp(x)\log(p(x)/q(x))$, with zero-mass terms set to zero
and a positive mass divided by zero giving infinity. Pinsker's inequality
$\operatorname{TV}(p,q)\le\sqrt{D(p\Vert q)/2}$, with natural logarithms
\citep[Lemma 2]{canonne2022kl}, applied conditionally, gives
\[
 \delta_\mu(F,E)
 \le\sqrt{\bigl(I_\mu(\Theta;E)-I_\mu(\Theta;F)\bigr)/2}.
\]
This bound uses conditional errors averaged over worlds, not the distance
between the two prior-predictive mixtures.

Fix a $\rho$-prefix $m$ and a $\pi$-prefix $n$. Average dominance provides
arbitrarily accurate decoders from sufficiently late $\pi$-histories into the
fixed $\rho$-history alphabet. Their output experiments are exact garblings of
the source and approximate $K_{\rho,m}$. Finite-output information is continuous
in average row total variation. Indeed, writing $E_\theta$ for an experiment's
row on the fixed finite output alphabet,
$I_\mu(\Theta;E)=\operatorname H(\int E_\theta\,d\mu)-\int\operatorname H(E_\theta)\,d\mu$.
This uses only finite-output entropy \citep[Sections 2.2--2.4]{cover2006elements}:
it is bounded and uniformly continuous on the finite simplex, so convergence
in mean row distance makes both terms converge. The world alphabet need not be finite.
Hence the decoded experiments' information approaches that of $K_{\rho,m}$.
Apply the reverse bound to each exact garbling, and then delete
the reconstructed source's suffix to time $n$. The triangle inequality, followed
by the approximation limit, yields
\[
 \delta_\mu(K_{\rho,m},K_{\pi,n})
 \le\sqrt{\bigl(J_I(\pi)-I_\mu(\Theta;K_{\rho,m})\bigr)/2}.
\]
Taking $m$ to infinity proves the displayed residual bound; deficiency is
also at most one. Equal totals therefore imply reverse average dominance.
Conversely, approximate simulation, data processing and fixed-output
continuity imply weak monotonicity of complete information. Applying it in
both directions gives equality.
\end{proof}

This equality condition gives an exact test for strictness in the desired
uniform order. On a specified feasible family, suppose first that every
information total is finite. Complete information is strictly finitarily
monotone exactly when
\[
 \pi\succeq_{\mathrm{fin}}\rho\quad\text{and}\quad
 \rho\succeq_\mu\pi
 \quad\Longrightarrow\quad
 \rho\succeq_{\mathrm{fin}}\pi.
 \tag{U}\label{eq:next-uniform-lift}
\]
To see necessity, a failure of this implication is a strict uniform pair that
is equivalent in the average order, so its information scores tie. For
sufficiency, any tied uniformly comparable pair is reverse-average comparable
by the theorem, and the implication removes strictness of the uniform pair.
With possibly infinite totals one must additionally exclude strict uniform
pairs whose two scores are both infinite. The finite-total theorem cannot
rule out such a tie.

\subsection{Positive guarantees for finite and compact classes}

The average-order conclusion is useful on its own. It also gives uniform
reliability when additional structure relates the two errors. For a finite
world class and a prior positive on every world, put
$a=\min_\theta\mu\{\theta\}>0$. Every common decoder's worst-world error is
at most its prior-average error divided by $a$. Thus average and uniform
finitary comparison agree, and information cannot tie a strict uniform pair.
The corresponding conclusion holds for a broader class of posterior scores.

\citet{garcia2012divergences} relate finite-hypothesis experiment comparison
to divergences and decision risks. The result below concerns growing histories
and the conversion from prior-average to uniform simulation.

Practical dynamics-information methods such as VIME approximate information
gain through variational Bayesian neural-network posteriors
\citep{houthooft2016vime}. The objectives here instead use a supplied world
class and exact posteriors.

\begin{proposition}[Finite-world posterior strictness]
\label{prop:next-finite-posterior}
Suppose the world class and action and observation sets are finite and nonempty,
and the prior $\alpha$ is positive on every world. Let $\Psi$ be continuous
and strictly convex on the probability simplex of worlds. If $p_{\pi,t}$ is
the posterior probability vector after time $t$, then
\[
 J_\Psi(\pi)=\lim_t
 \bigl[\mathbb E_\alpha\Psi(p_{\pi,t})-\Psi(\alpha)\bigr]
\]
is strictly finitarily monotone. Complete information and Brier returns are included. If a natively sufficient policy is
feasible, the exact maximizers are precisely the feasible sufficient policies.
Here \emph{full revelation} means an experiment that reports the world exactly.
Neither full revelation nor a natively sufficient policy is assumed attainable
for the strictness assertion itself.
\end{proposition}

\begin{proof}
An exact garbling makes the coarser posterior the conditional mean of the
finer posterior. Convexity therefore gives data processing. Continuity on
the compact simplex makes $\Psi$ bounded; the increasing expected potentials
have finite limits. Fixed-output score continuity and approximate simulation
then give weak finitary monotonicity.

The extra content of strict convexity is a uniform reverse modulus: sufficiently
small potential loss under an exact garbling allows sufficiently accurate
reconstruction of its source experiment. To make the reconstruction explicit,
let $X$ be the signal from experiment $E$ and $Y$ its garbled signal from $F$.
Write $P$ for the posterior given $X$ and $\bar P=\mathbb E[P\mid Y]$ for the
posterior given $Y$. The potential loss is
$\Delta=\mathbb E\Psi(P)-\mathbb E\Psi(\bar P)$.
The Bayes reverse decoder samples $X$ from its conditional law given $Y$
under the prior. Its average row error gives the bound
\[
 \delta_\alpha(F,E)\le\tfrac12\mathbb E\|P-\bar P\|_1.
\]
Thus it remains to show that small $\Delta$ forces small posterior distance.
Draw a conditionally independent copy $P'$ of $P$ given $Y$. Strict convexity
and compactness bound the midpoint convexity gap away from zero whenever
$P$ and $P'$ are separated by a fixed positive distance. Its expectation is
at most $\Delta$, so small loss forces $P$ and $P'$ to be close except on a
set of small probability. Since $\bar P$ is their conditional mean and
posterior distances are bounded, $\mathbb E\|P-\bar P\|_1$ tends to zero
with $\Delta$. This modulus depends only on $\Psi$ and the fixed world
simplex, not on the signal alphabet sizes. Dividing the average row error
by $\min_\theta\alpha(\theta)>0$ converts it to uniform row error.

If $\pi\succeq_{\mathrm{fin}}\rho$ but the reverse fails, one fixed
$\pi$-prefix remains at least $\varepsilon>0$ away from every $\rho$-prefix.
For any chosen $\rho$-prefix, use a long $\pi$-prefix to simulate it nearly
exactly. A nearly lossless simulation would, by the reverse modulus and score
continuity, reconstruct the fixed missing prefix to error less than
$\varepsilon$. Therefore every $\rho$-prefix has score at least one fixed
positive amount below the complete $\pi$-score. Taking the supremum preserves
this strict gap. The optimizer conclusion follows as in the native-score
proof.
\end{proof}

For the quadratic potential $\Psi(p)=\sum_\theta p(\theta)^2$, the increase
in expected potential over one additional observation equals the expected
squared change in the posterior. Summing this identity identifies the complete
potential difference with the complete Brier return. For negative
entropy, $\Psi(p)=\sum_\theta p(\theta)\log p(\theta)$, with $0\log0=0$,
the same potential difference is mutual information.

The finite-world result does not exhaust the positive cases. The next result
uses compactness in the world parameter space: a space is compact when every
open cover has a finite subcover. Continuity here means that every finite
controlled-prefix probability varies continuously with that parameter. A
Hausdorff space separates distinct points by disjoint open neighborhoods; its
Borel sigma algebra is generated by its open sets.

\begin{proposition}[A compact positive-atom guarantee]
\label{prop:next-compact-posterior}
Let the world parameter space be nonempty, compact, and Hausdorff, equipped with
its Borel sigma algebra. Suppose the controlled-behavior map is continuous and
a probability prior gives positive mass to every singleton world. Then the
prior-average and uniform finitary process preorders agree. Complete information strictly
respects their strict comparisons whenever the dominating total is finite;
finite prior entropy is sufficient for all these totals to be finite.
On a countable such class, complete Brier return for the posterior over all
worlds is strictly finitarily monotone without an entropy assumption.
\end{proposition}

\begin{proof}
Fix a finite target $T$ with output alphabet $Y$ and a growing source process
$E_t$ whose average deficiency tends to zero. Choose nearly optimal decoders.
Each world's error is bounded by the average error divided by that world's
positive prior mass, so it tends to zero.

To turn this pointwise convergence into uniform convergence, let $2^Y$ denote
the finite collection of subsets of $Y$, and let
$\Lambda=\mathcal P(\Q\times2^Y)$ be the probability distributions on
world--subset pairs $(\theta,B)$. Give $2^Y$ the discrete topology and
$\Lambda$ the weak topology, in which integrals of continuous functions
converge. This space is compact. For a fixed $\lambda\in\Lambda$, consider
the decision problem in which choosing $y\in Y$ earns $\mathbf1\{y\in B\}$.
The decision maker sees only the experiment's signal, not $(\theta,B)$.
Write $T(\theta,B)=\sum_{y\in B}T(\theta,y)$, let $x$ range over the source
signals, and define
\[
 \begin{aligned}
 R_T(\lambda)&=\int T(\theta,B)\,d\lambda(\theta,B),\\
 V_t(\lambda)&=\sum_x\max_{y\in Y}
       \int E_t(\theta,x)\mathbf1\{y\in B\}\,d\lambda(\theta,B).
 \end{aligned}
\]
Reporting the target's own output earns $R_T(\lambda)$;
$V_t(\lambda)$ is the best reward obtainable from the source. The finite-target
randomization criterion \citep[Proposition 6.2.3 and Theorem 6.4.1]{torgersen1991comparison},
with unit-range rewards and our total-variation normalization, gives
\[
 \delta(E_t,T)=\max_{\lambda\in\Lambda}
       [R_T(\lambda)-V_t(\lambda)]_+,
 \qquad [u]_+=\max\{u,0\}.
\]

The continuous finite experiment rows make $R_T$ and $V_t$ continuous in
$\lambda$, and prefix projection makes $V_t$ nondecreasing in $t$.
For each fixed $\lambda$, bounded convergence makes the chosen decoders'
rewards approach $R_T(\lambda)$, so
$[R_T(\lambda)-V_t(\lambda)]_+$ decreases to zero. Dini's theorem makes
this convergence uniform on compact $\Lambda$
\citep[Appendix M8, p.~242]{billingsley1999convergence}.
The displayed maximum therefore tends to zero. No uniform positive lower
bound on all prior atoms is required.
The two process orders therefore agree. Information strictness now follows
from Theorem~\ref{thm:next-information-equality}; the entropy ceiling supplies
finite totals when the prior entropy is finite.

For Brier, choose finite sets of worlds carrying nearly all the prior mass;
no entropy bound is needed.
Let $V(E)$ be the expected squared norm of the posterior over all worlds after
experiment $E$. If $F$ is an exact garbling of $E$, its Bayes reverse decoder
satisfies, for every finite set $B$ of worlds,
\[
 \delta_\mu(F,E)
 \le\mu(B^c)+\tfrac12\sqrt{|B|\,[V(E)-V(F)]}.
\]
On $B$, Cauchy--Schwarz controls posterior absolute differences by squared
movement; on its complement, error is at most one. Countability supplies
finite sets with arbitrarily small prior tail. Choosing the set first gives a
vanishing reverse modulus independent of the signal alphabets. The same
fixed-missing-prefix argument used above makes equal complete Brier values
equivalent to reverse average simulation. Compact lifting then gives strict
uniform comparison. The finite squared-movement identity telescopes, and its
nonnegative series converges to the bounded complete potential difference.
\end{proof}

Positive mass at every singleton is stronger than positive mass in every
nonempty open set. The latter, often called topological full support, is not
the hypothesis of this proposition. The conclusion is an eventual comparison
for each fixed target; it supplies no common acquisition deadline or method
for finding a sufficient policy.

\subsection{Why terminal information can erase finite access}

A terminal experiment has the entire infinite history as its signal. Two such
experiments are Blackwell equivalent when each can be simulated exactly by a
measurable world-independent decoder of the other. These decoders may inspect
the entire path; they need not finish at a finite time.

\begin{proposition}[Terminal equivalence and finite access]
\label{prop:next-terminal-obstruction}
There is a countable-world interface with two policies $\pi_Q$ and $\pi_W$ such that
$\pi_Q\succeq_{\mathrm{fin}}\pi_W$ and $\pi_W\not\succeq_{\mathrm{fin}}\pi_Q$, but their
terminal experiments are Blackwell equivalent. Every score invariant under
terminal Blackwell equivalence therefore ties this strict process pair.
\end{proposition}

\begin{proof}
Take worlds $\mathbb N_0\cup\{\infty\}$. At the first step, QUERY reports
whether the world is finite, while WAIT returns zero. At every later step,
actions leave an automatic pulse stream unchanged: world $k$ emits its one
pulse at step $k+2$, and world $\infty$ never pulses. Policy $\pi_Q$ queries at
startup and then always waits; $\pi_W$ always waits.

Replacing the initial action--observation pair in a $\pi_Q$-history with
$(\mathrm{WAIT},0)$ reproduces the $\pi_W$-history distribution in each world.
At every finite time, however, the $\pi_W$-history is identical in the
no-pulse world and in a finite world whose pulse lies after that time.
The query target separates those worlds. Any one decoder must therefore make
error at least $1/2$ in one of them. A fair guessed bit attains this bound.
Thus the reverse finite-access comparison fails.

The complete infinite pulse stream identifies the world: return its pulse
time, translated into $k$, or return $\infty$ if no pulse occurs. Each finite
label is determined by a finite prefix. Such events are called cylinder events;
the no-pulse event is a countable intersection of them. The decoder is measurable. Once it has the
world label, it can generate either policy's complete path. Hence the two
terminal experiments simulate each other exactly, despite their strict
finitary comparison.
\end{proof}

The assertion concerns scores of the terminal Blackwell equivalence class.
An arbitrary score of a timestamped infinite-path presentation may retain
prefix information and need not tie the pair. The proposition alone also
asserts a tied strict pair, not global optimality of every terminal-invariant
score over every policy. The shared alarm construction's global complete
information and Brier ties are separately established by their explicit
objective formulas.

\section{The alarm construction and specified objectives}
\label{app:alarm}

This appendix supplies the literal reward definitions behind the comparison.
All returns below are expectations under the same geometric prior,
$\alpha(\infty)=1/2$ and $\alpha(k)=2^{-(k+2)}$, except where a fixed-world
quantity is explicitly used before averaging. The native comparison takes a
supremum over worlds and does not use this prior. Logarithmic rewards are reported in bits.
For nonnegative rewards, the complete return is the limit of the finite returns. A discounted return includes the startup decision with weight one.
These aggregation conventions are part of each objective.

\subsection{The interface and its exact native error}
Use the interface defined in Section~\ref{sec:alarm}, abbreviating INSPECT,
PLAY$_0$, and PLAY$_1$ by $\mathsf I,\mathsf P_0,\mathsf P_1$. We retain every
action and observation in the history. For the calculations below, write
$b(\theta)=\mathbf1\{\theta<\infty\}$ and $g(a)=\mathbf1\{a=\mathsf P_1\}$;
the first is the inspection answer and the second determines the active display's
action-dependent digit.

Let $s=\pi(\mathsf I\mid\varnothing)$ and let $T_{\mathsf I}$ denote the
actual one-observation inspection experiment, including its independent color
(written $F$ in Section~\ref{sec:alarm}).
For a target experiment $T$, write $\delta(K_{\pi,t},T)$ for its deficiency
from the length-$t$ history acquired by $\pi$. The depth-$n$ native audit is the
supremum of this deficiency over the finite interventions of length at most
$n$ available on the interface.

\begin{theorem}[The common audit]\label{thm:next-alarm-audit}
For every randomized history policy $\pi$ and every $t\ge1$,
\[
 \delta(K_{\pi,t},T_{\mathsf I})=\frac{1-s}{2}.
\]
For every $t\ge n\ge1$, the depth-$n$ native audit has the same value.
Inspection simulates every policy's experiment at the same positive horizon.
Consequently $\pi$ is natively sufficient exactly when $s=1$; when $s<1$, an
inspection policy strictly dominates $\pi$ in the finitary process preorder.
\end{theorem}
\begin{proof}
Fix $t$ and compare $\infty$ with a finite world whose pulse comes after $t$.
Conditional on a play startup, their history laws agree, including the adaptive
action factors. Conditional on inspection, the root labels have disjoint
supports. Since the initial action remains in the history, the total variation
distance of the two source rows is exactly $s$. The corresponding target rows
have distance one. If a decoder has error at most $\varepsilon$ in each world,
contraction and the triangle inequality give $1\le s+2\varepsilon$.

For the upper bound, extract all alarm observations in the source history. An inspected
history also gives the correct bit $b$. On a play history, guess $b$ fairly.
Choose a world consistent with the bit and the recorded alarm prefix, and
simulate the target with its own policy randomness and fresh colors. If the
guessed bit is consistent and correct, all worlds still compatible with this
summary have identical target laws through depth $n\le t$. An impossible
summary can be filled arbitrarily. Thus inspection is exact, and the play
branch contributes error at most $1/2$. This proves the audit upper bound;
the startup target gives equality. The same decoder at $s=1$ proves
same-horizon sufficiency. A positive startup deficit at every source horizon
prevents reverse finitary domination.
\end{proof}

The lower bound concerns uniform simulation across worlds: the late-alarm world
may change with $t$, so the argument does not require one fixed finite alarm to
remain hidden forever. The upper bound likewise concerns simulated observations;
regenerating the display's random outputs from retained evidence does not make
the inspecting agent's physical display responsive again.

\subsection{Continuing pseudo-count bonuses at arbitrarily large horizons}
Let $N_h(o)$ count occurrences of observation label $o$ in a length-$t$ history $h$. A categorical probability model assigns one adjustable count
to each label. With initial count one on each of the four labels, its probability
before and after one further occurrence of $o$ is
\[
 \rho_h(o)=\frac{N_h(o)+1}{t+4},\qquad
 \rho'_h(o)=\frac{N_h(o)+2}{t+5}.
\]
The density-update construction of \citet[Equations (1)--(2)]{bellemare2016unifying}
defines the effective count by
\[
 \widehat N_h(o)=\frac{\rho_h(o)(1-\rho'_h(o))}{\rho'_h(o)-\rho_h(o)}
               =N_h(o)+1.
\]
We reward $1/\sqrt{\widehat N_h(o)}$ before updating the model. This is the
inverse-square-root specialization with unit scale and the stated initialization;
it does not use their learned image model or numerical bonus offset. With
$S(n)=\sum_{j=1}^n j^{-1/2}$ and $S(0)=0$, the cumulative reward on a length-$H$
history is
\[
 R_H=\sum_{o=0}^3 S(N_{H}(o)),\qquad J_H^{\rm pseudo}(\pi)=\mathbb E R_H.
\]
Expectation includes the prior, observations, and policy randomization.

\begin{proposition}[Continuing count bonus]\label{prop:alarm-pseudocount}
On the fixed alarm interface, for every integer $H\ge64$, the objective
$J_H^{\rm pseudo}$ has an attained global maximum over all randomized history
policies. Every maximizer skips initial inspection. Every extension of a maximizing
length-$H$ policy has inspection-target deficiency $1/2$ at every positive collection
time and eventually, and is strictly finitarily dominated by sufficient inspection.
\end{proposition}
\begin{proof}
The elementary bounds
\[
 2\sqrt{n+1}-2\le S(n)\le2\sqrt n,\qquad S(a+b)\le S(a)+S(b)
\]
follow by telescoping square-root increments and decreasing summands.
On any inspection branch there are at most two nonzero observations: the startup
label and the single possible alarm pulse. All other observations are zero,
regardless of subsequent actions. Hence every such branch satisfies
\[
 R_H\le S(H)+2\le2\sqrt H+2.
\]
By contrast, always choosing PLAY$_1$ gives $n=\lfloor H/2\rfloor$ display labels in
$\{2,3\}$ and at least $m-1$ zeros, where $m=\lfloor(H-1)/2\rfloor$.
For $H\ge64$, its reward on every supported history is at least
\[
 S(n)+S(m-1)\ge2\sqrt{n+1}+2\sqrt m-4
 \ge2\sqrt{2H-4}-4>2\sqrt H+2.
\]
These bounds survive prior expectation. Finite alphabets and a finite reward horizon
make the objective a finite multilinear function of policy rows, so a deterministic
finite plan attains its global maximum. Its best play branch strictly outperforms
every inspection branch. Splitting any policy by its initial action therefore shows
that a maximizer assigns zero probability to inspection. Changing its actions after
$H$ cannot change this conclusion. The native deficiency and strict domination follow
from Theorem~\ref{thm:next-alarm-audit}.
\end{proof}

Unlike finite returns at horizons $H\ge64$, the complete and long-run average
returns tie all policies. For every
length-$H$ observation sequence, subadditivity and the square-root bounds give
\[
 \sqrt H\le S(H)\le R_H\le2\sum_{o=0}^3\sqrt{N_H(o)}\le4\sqrt H.
\]
Thus $J_H^{\rm pseudo}(\pi)\to+\infty$ and
$J_H^{\rm pseudo}(\pi)/H\to0$ for every policy, including sufficient inspection.
Both limiting objectives tie all policies. This does not negate the proposition's
strict exclusion of inspection from every sufficiently long finite-horizon optimum.
The threshold $64$ is a convenient sufficient bound, not an optimal threshold.

\subsection{Information and Brier returns}
Write $\beta_t$ for the actual posterior over the countable world labels
given the complete length-$t$ history. The information objective at horizon $t$
is $I(\theta;H_t)$; its complete value is $\sup_t I(\theta;H_t)$.
The Brier return sums
$\mathbb E\|\beta_{t+1}-\beta_t\|_2^2$. It uses posterior probabilities
of individual worlds, without replacing the geometric tail by one label.
The name comes from the category-summed quadratic forecast loss of
\citet[Equations (2)--(4)]{brier1950verification}, whose optimal expected loss
at a distribution $p$ is $1-\|p\|_2^2$. Expected reductions in this optimal loss
give the posterior-movement identity used here:
\[
 \sum_{j<t}\mathbb E\|\beta_{j+1}-\beta_j\|_2^2
 =\mathbb E\|\beta_t\|_2^2-\|\alpha\|_2^2.
\]
Here $H(\alpha)=2$ bits and $\|\alpha\|_2^2=1/3$.
Every policy's history yields its alarm prefix by deterministic extraction.
The information of those prefixes converges to two bits and their posterior
quadratic potential converges to one. Data processing supplies lower bounds
for each policy; prior entropy and the unit quadratic-potential bound
supply the matching upper bounds. The convergent movement telescope therefore
gives, for every policy,
\begin{equation}\label{eq:next-complete-posterior}
 \sup_t I(\theta;H_t)=2,\qquad
 \sum_{t\ge0}\mathbb E\|\beta_{t+1}-\beta_t\|_2^2=\frac23.
\end{equation}
Both complete objectives have bad maximizing ties.

At each fixed positive finite horizon, however, inspection maximizes both
information and Brier returns, and equality holds exactly for
$s=1$. Theorem~\ref{thm:next-alarm-audit} supplies a same-horizon garbling
from inspection and rules out the reverse garbling when $s<1$. Strictness
of the actual full-support atomic information and Brier functionals gives
the strict preference. This positive result concerns the same countable
class; the change is from a complete return to a fixed finite horizon.

\subsection{Sensor empowerment}\label{app:alarm-empowerment}
Fix a world $\theta$, a visited history $h$, and a positive integer $n$.
A hypothetical probe chooses an action sequence $U=(A_1,\ldots,A_n)$ before seeing
any of the probe's observations, executes it through the response law, and keeps
only the final raw sensor reading $Y_n$. Its \emph{$n$-step sensor empowerment} is
\[
 C_{\theta,n}(h)=\max_{\nu\in\Delta(A^n)} I_\nu(U;Y_n),
\]
where $\Delta(A^n)$ is the set of probability distributions on length-$n$ action
sequences, and information is measured in bits. The world and starting history are
fixed in this channel. The probe defines a reward for a continuation from $h$;
our exploration targets still start from the empty history. The hypothetical
input distribution $\nu$ is optimized separately from the actual collecting policy. In particular, the output does not
include a copy of the actions.

At a nonempty history of length $t$, the last response in the probe comes from
the display exactly when $t+n-1$ is odd. Earlier probing actions cannot change
the startup mode or the alarm schedule. An alarm tick has an action-independent
output in the fixed world, as does a display tick in inspection mode. When the
probe ends at a display tick in play mode,
$Y_n=2B+C$, where $B$ is the final action's group and $C$ is a fresh fair color bit.
Thus $I(U;Y_n)=\operatorname H(B)\le1$; a fair choice between constant
PLAY$_0$ and PLAY$_1$ sequences attains one bit. Consequently,
\[
 C_{\theta,n}(h)=
 \mathbf1\{h\text{ did not start with inspection}\}
 \mathbf1\{|h|+n-1\text{ is odd}\},\qquad h\ne\varnothing.
\]

Let $\Theta$ have the paper's prior $\alpha$, and let $H_t$ be the collecting
policy's actual history after $t$ observations. For a reward horizon $N\ge1$,
define the finite return and the prior-averaged startup capacity by
\[
 S_{n,N}(\pi)=\mathbb E_{\alpha,\pi}
       \sum_{t=0}^{N-1}C_{\Theta,n}(H_t),\qquad
 c_n=\mathbb E_\alpha C_{\Theta,n}(\varnothing).
\]
The constant $c_n$ is finite and independent of the collecting policy. If that
policy inspects initially with probability $s$, its display remains active with
probability $1-s$ at every later history, regardless of subsequent actions. Counting
the positive starting times whose probe ends at a display tick gives
\[
 S_{n,N}(\pi)=c_n+(1-s)m_{n,N},\qquad
 m_{n,N}=\begin{cases}
 \lfloor N/2\rfloor,&n\text{ odd},\\
 \lfloor(N-1)/2\rfloor,&n\text{ even}.
 \end{cases}
\]
For every fixed $n\ge1$, both counts divided by $N$ converge to $1/2$ along the
full sequence of reward horizons. Hence the long-run average return is
\[
 \overline J_{{\rm emp},n}(\pi)
   =\lim_{N\to\infty}\frac{S_{n,N}(\pi)}{N}
   =\frac{1-s}{2}.
\]
All and only policies with $s=0$ maximize this score, so maxima are attained with
arbitrary subsequent continuation. Every maximizer has inspection deficiency $1/2$
at every positive collection time and in the eventual infimum. Sufficient
inspection strictly finitarily dominates each one. The same interface and optimizer
set work for every positive probe horizon; increasing $n$ changes only the phase
of the alternating endpoint capacities.

The complete return does not give this comparison: whenever
$s<1$, the finite returns above grow without bound. Dividing by the reward
horizon compares the resulting rates. Thus the long-run average return is a specified temporal
objective, justified here by divergence of the total, rather than part of the
channel-capacity definition itself. This distinction matters: the complete
return has more maximizing policies.

The same failure holds under every proper discount on the same interface.
For $0<\gamma<1$, the discounted visited-capacity return is
\begin{equation}\label{eq:alarm-multistep-discounted-empowerment}
 \begin{aligned}
 J_{{\rm emp},n}^{\gamma}(\pi)
 &=\mathbb E_{\alpha,\pi}\sum_{t\ge0}\gamma^tC_{\Theta,n}(H_t)
   =c_n+(1-s)a_{\gamma,n},\\
 a_{\gamma,n}&=\begin{cases}
 \gamma/(1-\gamma^2),&n\text{ odd},\\
 \gamma^2/(1-\gamma^2),&n\text{ even}.
 \end{cases}
 \end{aligned}
\end{equation}
Indeed, pair the consecutive positive times $2k+1,2k+2$. Only one term
in each pair can have positive display capacity, and the geometric series sums to the
stated coefficient. The coefficient is positive, so all and only $s=0$ policies
maximize the discounted return, with the same permanent deficiency $1/2$.
This keeps the sensor, probe, collecting policies, prior, and environment fixed;
only the aggregation of visited capacities changes.

\citet[Section~4.3]{mohamed2015variational} explicitly discuss long-term,
potentially discounted planning with empowerment as a reward. That passage
supports this return convention; it does not identify the exact physical-channel
calculation here with their learned estimator or experimental optimization procedure.
Both comparisons use open-loop probes and the final raw sensor only. They do
not cover full-observation-block empowerment or probes that adapt to
intermediate sensors.

\subsection{The native score on this interface}
Inspection is sufficient by Theorem~\ref{thm:next-alarm-audit}.
For the rich target family and strictly positive weights of
Theorem~\ref{thm:strict-finitary-objective}, the maximizing policies are therefore
exactly the sufficient policies:
\[
 \pi\in\operatorname*{arg\,max}J_w\quad\Longleftrightarrow\quad s=1.
\]

\section{Compression progress and acquired evidence}
\label{app:compression}

Compression progress rewards improvement in describing the stored data, or
\emph{archive}. The old and updated compressors are assessed on the same
current archive, so their difference measures the effect of the update
\citep[Section II.A, Equation (2)]{schmidhuber2010formal}.
The ideal full-history example below uses complete returns on a separate binary
interface; its target is not the alarm inspection experiment. Its scope is a
fixed history scorer: the description language and all background supplied to
the scorer are fixed before choosing which of two buttons reveals the unknown
world. It therefore does not compare compressors separately informed of each
possible-world class.

\subsection{The ideal learner and its return}
Fix an \emph{interpreter} $U$: an algorithm that runs a finite binary program
with a finite binary auxiliary input. It may fail to halt; a successful run
outputs a finite binary string. A \emph{description} of some data is a program
that reconstructs those data. We call the interpreter \emph{complete} if every
finite string has a description at every auxiliary input. Require this property,
and write $K_U(x\mid y)$ for the shortest program length producing $x$ given $y$. For prefix Kolmogorov complexity, no successful
description is a prefix of another at the same auxiliary input, and its shortest descriptions are no longer than those of any other complete
prefix interpreter, up to a constant depending only on that interpreter.
These properties are called \emph{prefix-freeness} and \emph{optimality}
\citep[Appendix A, Definition 11]{vitanyi2000minimum}.

A compression model $m$ is a finite binary string used as auxiliary information. It may
describe a predictor or memorized data, but need not be a probability
distribution; every finite string is allowed. On the binary interface below,
encode the full history $h=((a_1,o_1),\ldots,(a_t,o_t))$ by
$D(h)=a_1o_1\cdots a_to_t$. Write $\epsilon$ for the empty string. Following
the model-plus-data form of ideal minimum description length
\citep[Section II, Definitions 1--2]{vitanyi2000minimum}, our cost is
\[
 C_U(m,h)=K_U(m\mid\epsilon)+K_U(D(h)\mid m).
\]
The first term charges for the model and the second for the residual program.
This two-part cost is not being identified with one-part $K_U(D(h)\mid\epsilon)$.
The cited compression-progress principle allows different compressors and
background knowledge. Our specified two-part oracle learner is a mathematical
instance of a same-archive update score; the theorem below does not claim a
source-wide failure of compression or a comparison with equal class-specific
background information.

Start from a fixed model $m_0$. After observing $h_t$, choose a model minimizing
$C_U(m,h_t)$ over all finite binary strings, retaining $m_{t-1}$ whenever it
is optimal and otherwise using a fixed archive-dependent choice. A minimum
exists because these costs are natural numbers and the model class is nonempty.
The update reward and cumulative history score are
\[
 r_t=C_U(m_{t-1},h_t)-C_U(m_t,h_t),\qquad
 F_H(h_H)=\sum_{t=1}^{H}r_t.
\]
Every gain is a finite nonnegative integer. These are exact oracle updates with
unlimited computation, full archives, and no model-size or codebook restriction.

\subsection{A two-button example and a complete optimum}
There are two worlds, indexed by a bit $q$, with the fair prior. Actions and
observations are also bits. A \emph{wiring} $e\in\{0,1\}$ specifies which button
reveals $q$ on the first press; the other button returns one independent fair
bit $B$. Every later observation repeats the first observation, regardless of
later actions. The bit $B$ is drawn once and is response randomness, not another
unknown world parameter.

The interpreter, initialization, encoding, and selection rule remain fixed
across wirings. Any \emph{background information}, meaning information supplied
to the scorer beyond its collected history, must also remain fixed. In particular,
the scorer is not separately told which button reveals the world.
\citet{looks2010compression} describes Rayhawk's pseudorandomness example, in which
a computationally limited compressor gains by obtaining a random seed. That
criticism concerns subjective interestingness and depends on computational cost;
the example below concerns native evidence and permits exact shortest descriptions.

Let $R_e$ always press $e$ and let $T_e=K^e_{R_e,1}$ be its one-step history
experiment. This policy reveals the whole world and can simulate every finite
native target. A policy that initially chooses $1-e$ has a full history independent
of $q$ forever, even with adaptive later actions, and
\[
 \delta(K^e_{\pi,t},T_e)=\tfrac12\quad\text{for every finite }t\ge0.
\]
For the lower bound, every decoded answer has the same distribution in the two
worlds, whereas $T_e$ returns opposite world bits. Their two errors cannot both
be below $1/2$; a fair answer gives the matching upper bound.

For every fixed policy and positive-length history, the prior-averaged law is
\begin{equation}\label{eq:compression-swap-law}
 \overline P_e^\pi(h)
 =\tfrac12\prod_{i=1}^{|h|}\pi(a_i\mid h_{i-1})
   \mathbf1\{o_1=\cdots=o_{|h|}\}.
\end{equation}
It does not depend on $e$. Define the finite return and its complete
undiscounted limit by
\[
 J_{e,H}(\pi)=\mathbb E_{e,\pi}F_H(H_H),\qquad
 J_e(\pi)=\sup_{H\ge0}J_{e,H}(\pi)\in[0,\infty].
\]
The expectation averages the fair world prior, response randomness, and policy
choices. Nonnegative gains make the complete return well-defined, possibly infinite.
Every fixed learner has the same objective for a given policy in either wiring.
This equality retains action labels and does not assume that their Kolmogorov
complexities are invariant under renaming.

\begin{theorem}[A dominated complete compression optimum]
\label{thm:complete-compression-optimum}
For every fixed ideal learner above, one of the two wirings has an attained
global maximizer $\pi_*$ of complete return over all randomized history policies,
with
\[
 R_e\succ_{\rm fin}\pi_*,
 \qquad
 \delta(K^e_{\pi_*,t},T_e)=\tfrac12\quad(t\ge0).
\]
If complete scores have a finite upper bound across policies, the maximizing
value is finite. Without that bound, the maximum may be infinite. The same
conclusion holds for any fixed exact learner with a nonempty model class
and natural-valued same-archive costs.
\end{theorem}
\begin{proof}
Let $M_H=\max_\pi J_{e,H}(\pi)$ and $M=\sup_\pi J_e(\pi)$.
At each finite horizon, every randomized policy's score is a convex combination
of scores of finitely many deterministic contingent plans
\citep[Lemma 3]{kuhn1953extensive}, so a deterministic plan attains $M_H$.
For $H\ge1$, its two possible observation paths have probabilities $1/2$ and
integer gains, while $M_0=0$. Consequently $M_H\in\frac12\mathbb N$, and
$M=\sup_HM_H$.

If $M<\infty$, the nondecreasing sequence $M_H$ reaches $M$ at some finite
$H_*\ge1$. Extend a deterministic maximizing plan arbitrarily. Its complete
score is at least $M_{H_*}$ by nonnegativity and at most $M$ by definition,
so it maximizes complete return. Choose the wiring opposite its first action.
Equation~\eqref{eq:compression-swap-law} preserves optimality, while the
world-independent branch gives the stated deficiency and strict domination.

If $M=\infty$, conditioning on the initial action shows that one fixed-root
policy family has unbounded complete scores. Choose policies $\pi_n$ in that
family with $J(\pi_n)\ge n2^n$, and mix them with weights $2^{-n}$ for $n\ge1$.
This is realizable as a history policy: writing
$w_n(h)=\prod_{i\le |h|}\pi_n(a_i\mid h_{i-1})$, use
\[
 \pi(a\mid h)=
 \frac{\sum_{n\ge1}2^{-n}w_n(h)\pi_n(a\mid h)}
      {\sum_{n\ge1}2^{-n}w_n(h)}
\]
when the denominator is positive and any fixed distribution otherwise.
Only action-probability factors appear in these weights: at any history
possible in a fixed world, the same world-response likelihood multiplies
every component and cancels when the mixture weights are normalized.
Its history laws are the corresponding mixtures in every world. For each $n$,
nonnegative rewards give $J(\pi)\ge2^{-n}J(\pi_n)\ge n$, so its expected
complete return is infinite. Its initial action is the
common root action; select the opposite wiring as before.
\end{proof}

The learner and background are fixed before the wiring, and the conclusion gives
one deficient optimum, not failure of every optimum or of one wiring for every
learner. Exact information gain about $q$, supplied with the actual wiring,
instead gives one bit for revealing and zero for the independent bit; it does
not satisfy the shared-scorer hypothesis. Supplying that wiring to a compressor
can likewise change its history score and invalidate the cross-wiring equality.
The obstruction is that one fixed scorer cannot universally rank histories by
which worlds they distinguish.

\subsection{Finite values and the description language}
Finiteness cannot be inferred from one bit of world uncertainty. The archive
keeps growing, and the rewards compare different archives at different updates,
so their sum need not telescope. The following constructions show both possibilities
for optimal prefix interpreters while keeping all finite models available.

Fix a complete optimal prefix interpreter $V$, abbreviate
$K_V(x)=K_V(x\mid\epsilon)$, and choose a constant $c$ such that
\[
 K_V(D)\le K_V(m)+K_V(D\mid m)+c.
\]
Concatenating the two self-delimiting descriptions and applying optimality gives
this bound. Fix $L\ge\max\{2,c+2\}$. An instruction of the form $1^Lp$ means
$L$ ones followed by program $p$; it simulates $V$ with the same auxiliary input.

First let the one-bit program 0 copy the auxiliary input only when it is
nonempty, and let the simulation instruction be the only other halting form.
This defines a complete optimal prefix interpreter $U$. For every nonempty
history $h$, put $d=D(h)$. Model $d$ costs $L+K_V(d)+1$, while every other
model costs at least $2L+K_V(d)-c$, which is strictly larger. The unique optimum is therefore the
current archive itself. Each update after the first earns at least one unit,
so every policy's complete return is infinite.

For a finite-valued example, instead let program 0 output $\epsilon$ and program
10 copy the auxiliary input, retaining the simulation instruction. These three
forms are prefix-free, and universal simulation preserves optimality.
For every nonempty history $h$, again put $d=D(h)$. Then
\[
 C_U(\epsilon,h)=L+K_V(d)+1,\qquad
 C_U(d,h)=L+K_V(d)+2,
\]
while any other nonempty model costs at least $2L+K_V(d)-c$.
Thus $\epsilon$ uniquely minimizes cost for every nonempty archive. From any
nonempty initialization the first gain is finite and positive, and every later
gain vanishes. There are only four one-step histories, so all policies' complete
scores have a uniform finite bound. The theorem supplies a strictly dominated
complete global maximizer with a finite positive value.

This last example establishes that infinite scores and a limited codebook are
unnecessary for the obstruction. Its description language is deliberately
chosen, however, and the reward comes entirely from correcting initialization.
The general theorem permits every fixed interpreter; its finite-valued
specialization cannot be promoted to a finiteness guarantee for all interpreters.

\section{Discounted predictive Shannon and Square rewards}
\label{app:predictive}

We use the unchanged alarm interface and geometric prior of
Section~\ref{sec:alarm}. These rewards score the next raw observation, whereas
the posterior rewards in the main text score changes in beliefs about worlds.
Let $q(o\mid h,a)$ be the Bayesian probability of next observation $o$ after
history $h$ and proposed action $a$. It averages the world's response probabilities
under the posterior given the complete history. The source's per-step Shannon and
Square utilities are, respectively,
\begin{equation}\label{eq:predictive-utilities}
 r^{\rm Sh}(h,a,o)=-\log_2 q(o\mid h,a),\qquad
 r^{\rm Sq}(h,a,o)=-q(o\mid h,a).
\end{equation}
These are the one-step definitions in
\citet[Section~2.3.2, Definitions~16--17, Equations~(2.15)--(2.16)]{aslanides2016thesis}.
This per-step Square utility differs from the whole-history Square-KSA in
\citet[Section~5]{orseau2013universal}; the result below concerns the thesis's
utility.
For $0<\gamma<1$, define the discounted returns
\[
 J^X_\gamma(\pi)=\sum_{t\ge0}\gamma^t
   \mathbb E_\pi[r^X(H_t,A_{t+1},O_{t+1})],
 \qquad X\in\{\mathrm{Sh},\mathrm{Sq}\}.
\]
The expectation includes the same prior, environmental randomness, and arbitrary
randomized history policy as the other Bayesian alarm objectives. Startup has
weight one. The source permits Bayesian mixtures over countable classes of possible worlds,
history-dependent randomized policies, and geometric discounts
\citep[Sections~2.2.1--2.2.2, Equation~(2.10), and Definition~14]{aslanides2016thesis}.

\paragraph{The source's environment assumptions.}
The thesis assumes finitely many actions, percepts, and hidden states in each
world's hidden-state realization, with a stochastic state transition followed by a percept
emitted from the new state. It does not require the percepts to be Markov or
the hidden-state realizations of different worlds to have a common finite size.
Every alarm world has such a realization. Its hidden state stores the startup
mode, the display/alarm phase, and a countdown to its pulse; after the pulse,
the countdown remains zero. The never-pulse world has countdown zero throughout.
A transition also stores the sampled raw label, which the new state emits
deterministically. This produces exactly the original response law at every
history and hence exactly the same history experiment under every policy.
None of these hidden coordinates is supplied to the agent. If a percept includes
an extrinsic reward coordinate, set that coordinate constantly to zero.

\paragraph{Finiteness of discounted returns.}
The expected Shannon reward at a history is the entropy of the forecast,
bounded by $\log_2 4=2$. Null outcomes contribute zero to this expectation.
The realized logarithmic utility need not be bounded; this is also noted after
the source's Shannon definition. Here its nonnegative discounted return
is finite. Square lies in $[-1,0]$ and is absolutely summable after discounting.
For a label $o$, let $\mathbf e_o$ denote the vector with one in coordinate
$o$ and zeros elsewhere. Square's relation to categorical squared prediction
error is the expectation identity:
\[
 \mathbb E_{O\sim q}[-q(O)]=-\sum_o q(o)^2
 =\mathbb E_{O\sim q}\|\mathbf e_O-q\|_2^2-1.
\]
Thus the two discounted returns differ by $1/(1-\gamma)$, independent
of policy. We use the literal utility in~\eqref{eq:predictive-utilities}; we do
not subtract divergent undiscounted totals.

\begin{proposition}[Predictive rewards on the fixed alarm]
\label{prop:alarm-predictive-discount}
For every $0<\gamma<1$, both $J^{\rm Sh}_\gamma$ and $J^{\rm Sq}_\gamma$ attain
their global maximum over all randomized history policies. For either objective,
a policy is maximizing if and only if it skips initial inspection with
probability one. Every maximizer has inspection-target deficiency $1/2$ at
every positive collection time and eventually, and is strictly dominated in
the finitary process preorder by sufficient inspection.
\end{proposition}
\begin{proof}
Write $s=\pi(\mathsf I\mid\varnothing)$. At alarm tick $k\ge0$, put
\[
 a_k=2^{-(k+2)},\qquad T_k=2a_k,\qquad
 S_k=\tfrac12+2a_k,\qquad p_k=a_k/S_k.
\]
Here $a_k$ is the prior mass of the world whose pulse occurs at that tick,
$T_k$ is the prior mass of finite worlds with no earlier pulse, and $S_k$
also includes the never-pulse world. On an inspected history with no earlier
pulse and a positive startup answer, the next pulse probability is $1/2$.
On a play history with no earlier pulse it is $p_k$. Already resolved histories
have deterministic forecasts. Let
$h_2(p)=-p\log_2 p-(1-p)\log_2(1-p)$, with zero endpoint terms.
The resulting expected stage utilities, conditional on the startup choice,
are
\[
\begin{array}{c|cc|cc}
 &\multicolumn{2}{c|}{\text{Shannon}}&\multicolumn{2}{c}{\text{Square}}\\
 \text{stage}&\text{inspect}&\text{play}&\text{inspect}&\text{play}\\ \hline
 t=0&2&2&-1/4&-1/4\\
 t=2k+1&0&1&-1&-1/2\\
 t=2k+2&T_k&S_kh_2(p_k)&-1+a_k&-1+2a_k(1-p_k).
\end{array}
\]
These expressions hold for all randomized adaptive continuations. At a display tick,
the chosen action fixes the digit and leaves exactly one fair color bit on a
play branch. At an alarm tick, the likelihood of the retained actions and display
labels cancels from the posterior ratio. Thus the expected stage reward of an
arbitrary policy is $s$ times its inspection column plus $1-s$ times its play
column.

Since $0\le p_k\le1/2$, binary-entropy concavity gives
$h_2(p_k)\ge2p_k$, so $S_kh_2(p_k)\ge T_k$.
Also $2a_k(1-p_k)\ge a_k$. Play therefore weakly improves every stage of
either score. The first display tick gives a strict improvement whenever $s>0$;
its discount weight is $\gamma>0$. Absolute convergence of the expected
series then shows that precisely the policies with $s=0$ maximize either
return. Any fixed play policy attains both maxima. The deficiency and strict
domination statements follow from Theorem~\ref{thm:next-alarm-audit}.
\end{proof}

The statement holds for every proper discount on this one environment. In
particular, increasing the discount toward one does not restore inspection:
the maximizing startup remains the same throughout. Attraction to stochastic
observations was already demonstrated for Shannon-KSA and whole-history Square-KSA
by \citet[Section~5]{orseau2013universal} and discussed for the per-step rewards
by \citet[Section~4.1.1]{aslanides2016thesis}. The conclusion here quantifies the
inspection deficiency of all exact objective optima; it makes no claim about
the thesis's approximate planning implementation or about other prediction-error
or learning-progress rewards.

\section{Posterior movement: an alarm success and a finite failure}
\label{app:posterior-movement}

\subsection{The published rewards and their aggregation}
Let $q=\beta_{t-1}$ and $p=\beta_t$ be the posterior over worlds before and after
one interaction. The Hellinger and Absolute knowledge-seeking rewards of
\citet[Section~3, equations~(3)--(5)]{orseau2013universal} are
\[
 r_{\rm Hel}(q,p)=\sum_\theta(\sqrt{p_\theta}-\sqrt{q_\theta})^2,
 \qquad r_{\rm Abs}(q,p)=\sum_\theta|p_\theta-q_\theta|.
\]
Neither expression has a factor $1/2$. The source permits general history-dependent
response kernels, a countable class of possible worlds with positive prior weights,
and both summable discount vectors and the all-ones vector. We use the same
variables and information access, with geometric weights $\gamma^{t-1}$ for
$0<\gamma<1$ or weights one for the complete return. These are the source's unnormalized discounted values; the remaining-discount
normalization in its convergence theorems is not part of the objective. Our calculations cover
randomized history policies, including the deterministic policies used by the
source's main statements.

For a retained history $h$ and chosen action $a$, write
$q_\theta(o)=Q_\theta(o\mid h,a)$ and
$\bar q(o)=\sum_\theta\beta_h(\theta)q_\theta(o)$.
Bayes' rule gives the equivalent conditional expected reward
\begin{equation}\label{eq:posterior-movement-predictive}
 \mathbb E[r_d(\beta_h,\beta_{haO})\mid h,a]
 =\sum_\theta\beta_h(\theta)\sum_o d(q_\theta(o),\bar q(o)),
 \quad d(x,y)=(\sqrt{x}-\sqrt{y})^2\ \text{or}\ |x-y|.
\end{equation}
Indeed, the new posterior coordinate is
$\beta_h(\theta)q_\theta(o)/\bar q(o)$, and multiplication by $\bar q(o)$
cancels the common scaling in each distance. Null observations contribute zero.
This identifies the source's expected reward on the full-history Bayesian
posterior. Squared posterior changes do not satisfy this scaling identity.

For a deterministic yes/no query whose posterior probability of YES is $p$,
\begin{equation}\label{eq:binary-posterior-movement}
 g_{\rm Hel}(p)=2\{1-p\sqrt p-(1-p)\sqrt{1-p}\},
 \qquad g_{\rm Abs}(p)=4p(1-p).
\end{equation}
A world-independent random observation pays zero to either reward. Put
$c=2-\sqrt2$; a fair binary query pays $c$ and $1$, respectively.

\subsection{Both rewards succeed on the shared alarm}
Retain exactly the alarm prior and interface: inspection initially reveals whether
an alarm will ever occur, while PLAY initially gives only a world-independent
random label. Alarm tick $k$, at observation $2k+3$, reveals whether the alarm rings then.
The interleaved display observations give only known color randomness in play mode and a
constant in inspection mode. Later actions cannot change that initial mode.
The never-alarm world has prior mass $1/2$. Put
\[
 a_k=2^{-(k+2)},\qquad T_k=2a_k,\qquad S_k=\tfrac12+2a_k,
 \qquad p_k=a_k/S_k.
\]
Under the original prior, finite worlds whose pulse has not yet occurred have
total mass $T_k$. Conditional on a positive inspection answer and no earlier
pulse, the next alarm observation is a fair query. Without inspection, the unresolved
event also includes the never-alarm world, giving prior mass $S_k$ and
conditional pulse probability $p_k$.
Previously observed pulses leave no world uncertainty. Conditioning on the
entire retained history cancels the known action and display likelihoods.
Consequently every policy has return
$sJ_\gamma(I)+(1-s)J_\gamma(P)$, with
\begin{align*}
 J_\gamma^{\rm Hel}(I)&=c+\sum_{k\ge0}\gamma^{2k+2}cT_k,&
 J_\gamma^{\rm Hel}(P)&=\sum_{k\ge0}\gamma^{2k+2}S_kg_{\rm Hel}(p_k),\\
 J_\gamma^{\rm Abs}(I)&=1+\sum_{k\ge0}\gamma^{2k+2}T_k,&
 J_\gamma^{\rm Abs}(P)&=\sum_{k\ge0}\gamma^{2k+2}4a_k(1-p_k).
\end{align*}
These series converge also at $\gamma=1$.

\begin{proposition}[Alarm posterior-movement success]
\label{prop:alarm-posterior-movement}
For either published reward and every $0<\gamma\le1$, all maximizing alarm
policies have $s=1$ and zero eventual inspection deficiency. Here $\gamma=1$
denotes the complete return.
\end{proposition}
\begin{proof}
It suffices to bound PLAY's extra rewards at alarm ticks by less than INSPECT's
startup payment. For Absolute the extra payment is $p_k$, and
\[
 0\le p_k\le 2a_k-\tfrac14\mathbf1_{k=0},
 \qquad \sum_kp_k\le\tfrac34<1.
\]
For Hellinger, $g_{\rm Hel}(p)\le3p-2p\sqrt p$ for $0\le p\le1$.
To verify the inequality, set $x=\sqrt{1-p}$ and use
$2x^3-3x^2+1=(x-1)^2(2x+1)\ge0$.
Since $S_k\le1$, the extra payment is bounded above by
\[
 u_k=(3-2c)a_k-\tfrac14\mathbf1_{k=0}
                     -\tfrac{\sqrt2}{16}\mathbf1_{k=1}\ge0.
\]
The two subtractions use $2a_0\sqrt{a_0}=1/4$ and
$2a_1\sqrt{a_1}=\sqrt2/16$; at every other index we discard a nonnegative
subtraction. Thus
\[
 \sum_ku_k=\frac{3-2c}{2}-\frac14-\frac{\sqrt2}{16}<c,
 \qquad c-\sum_ku_k=\frac{44-31\sqrt2}{16}>0.
\]
Multiplying each extra payment by a weight in $[0,1]$ preserves these upper
bounds, even if an extra payment is negative. Both return gaps are strictly
positive. The affine policy formula and the alarm deficiency identity finish
the proof. The formal proof includes the posterior normalizers, null-history
cases, series convergence, and all randomized continuation policies.
\end{proof}

\begin{proposition}[Proper-discount posterior controls]
\label{prop:alarm-discount-posterior}
On the same alarm, every maximizing policy for geometrically discounted
information gain or Brier rewards on the posterior over worlds has $s=1$
for every fixed $0<\gamma<1$. Their complete undiscounted objectives still tie
all policies, at two bits and $2/3$, respectively.
\end{proposition}
\begin{proof}
For this alarm, the expected successive-posterior information reward equals
$V_{t+1}-V_t$, where $V_t=I(Q;H_t)$ is measured in bits. The countable posterior
KL series is absolutely convergent on every positive-probability history.
For a bounded prefix value $V_t$, Abel summation gives
\[
 \sum_{t\ge0}\gamma^t(V_{t+1}-V_t)
   =(1-\gamma)\sum_{t\ge0}\gamma^tV_{t+1}-V_0.
\]
Use prefix information or the expected squared posterior norm. Inspection
weakly maximizes every prefix value, and strictly improves every positive
prefix when $s<1$. The common $V_0$ cancels and all weights are positive.
If $s=1$, all prefix values agree with inspection. This proves the stated
optimizer sets without interchanging optimization and $\gamma\uparrow1$.
\end{proof}

\subsection{Three incremental bits can outscore four bits at once}
The separate environment has sixteen worlds, the four-bit strings
$\theta=(B_1,B_2,B_3,B_4)$, with a uniform prior. The initial FULL action reveals
$\theta$ in one of sixteen observation labels. The initial PARTIAL action
reveals $B_1$, followed automatically by $B_2$ and $B_3$ at the next two
observations. All remaining observations are the same constant. Both actions
remain legal; actions after the first do not change the mode or schedule.
Each world has a deterministic finite-state realization. Let $F$ be the native
one-step FULL experiment, and let $s$ now denote initial FULL probability.

A uniform posterior on $m$ labels, split into two equal cells, earns one bit
of information, $1/m$ Brier movement, $c$ Hellinger movement, or one Absolute
movement. Revealing all sixteen labels immediately instead earns
$4$, $15/16$, $3/2$, or $15/8$, respectively. Therefore
\begin{center}
\begin{tabular}{@{}lll@{}}
\toprule
Objective & FULL return & PARTIAL return \\
\midrule
Information, bits & $4$ & $1+\gamma+\gamma^2$ \\
Ordinary Brier & $15/16$ & $1/16+\gamma/8+\gamma^2/4$ \\
Hellinger & $3/2$ & $(2-\sqrt2)(1+\gamma+\gamma^2)$ \\
Absolute & $15/8$ & $1+\gamma+\gamma^2$ \\
\bottomrule
\end{tabular}
\end{center}
Every policy's return is the initial mixture of these two values; there is
no restriction on its later randomization.

\begin{proposition}[Finite posterior-movement failure]
\label{prop:four-bit-posterior-movement}
For Hellinger and Absolute, every optimum has $s=0$ for
$9/10\le\gamma\le1$, and
\[
 \delta(K_{\pi,t},F)=\tfrac12\qquad(t\ge3).
\]
Information and ordinary Brier instead have exactly the policies with $s=1$
as optima for every $0<\gamma\le1$, with zero eventual deficiency.
The native score $J_w$ also has exactly the sufficient policies
as optima, with zero eventual deficiency.
\end{proposition}
\begin{proof}
The displayed returns give the strict comparisons. The conservative common
range $\gamma\ge9/10$ makes both Hellinger and Absolute strictly prefer PARTIAL;
$\gamma=1$ is the complete return. Information and Brier strictly prefer FULL
throughout $0<\gamma\le1$.

For every policy and $t\ge3$, the exact deficiency is $(1-s)/2$.
A decoder uses the revealed world on FULL histories and the three observed bits
with an independent fair guess for $B_4$ on PARTIAL histories, giving the upper
bound. For the lower bound, pair two worlds with the same first three bits and
opposite fourth bit. Their full-history laws coincide on PARTIAL histories under
any continuation policy. A common decoder's two probabilities of correctly
reporting these worlds sum to at most one on that branch; the FULL branch has
probability $s$ in either world. One row error is therefore at least $(1-s)/2$.
This argument holds at every later horizon. Finally, FULL reveals the world
and can simulate every finite native target. Theorem~\ref{thm:strict-finitary-objective}
therefore supplies the $J_w$ statement for the same feasible policy family.
\end{proof}

The comparison uses different paths of the same Bayesian posterior, not
different variables or learners. Expected information and Brier movement
are increments of fixed posterior potentials and telescope. Accumulated
Hellinger and Absolute movement also rewards how the posterior changes along
the path. Their alarm success and four-bit failure must therefore be read
together. The construction does not contradict the source's continuation
prediction guarantee: after PARTIAL, no available continuation reveals $B_4$;
the missed native experiment was the alternative initial FULL action.

\section{MOP in a fully observed finite MDP}\label{app:mop-mdp}
The maximum occupancy principle (MOP) maximizes discounted action entropy and
successor-state transition entropy in a finite time-homogeneous MDP
\citep[Equations (1)--(2) and Methods]{ramirezruiz2024occupancy}.
We use its stationary-policy objective with both entropy coefficients equal to
one. Entropies in this appendix use natural logarithms. The example and
native-deficiency calculation below are ours; the MDP objective is the source's.

\paragraph{States, actions, and the objective.}
There are five observed MDP states: START, $d_0,d_1,n_0,n_1$.
The initial state is START. An unknown bit $\theta\in\{0,1\}$ specifies the
world's transition law. At START, READ moves deterministically to $d_\theta$;
NOISE moves to $n_0$ or $n_1$ with equal probability, independently of
$\theta$. Each of the four remaining states is absorbing and has only one
available action, STAY. This follows the source's absorbing-state convention.
The agent observes its MDP state; no hidden clock or full-history state
is used. The unknown bit specifies the transition law, rather than a hidden
component of the state.

Write $P_\theta(\cdot\mid x,a)$ for this transition distribution and
$\mathcal A(x)$ for the available actions at $x$. A stationary policy
$\pi(\cdot\mid x)$ is a distribution on $\mathcal A(x)$. In each fixed MDP,
starting from $S_0=\mathrm{START}$, its objective is
\[
 J_{\gamma,\theta}^{\rm MOP}(\pi)=
 \sum_{t\ge0}\gamma^t\mathbb E_\theta^\pi\!\left[
 H(\pi(\cdot\mid S_t))+
 \sum_{a\in\mathcal A(S_t)}\pi(a\mid S_t)
 H(P_\theta(\cdot\mid S_t,a))\right],\qquad 0<\gamma<1.
\]
For selection without a world prior, define
$J_\gamma^{\rm MOP}(\pi)=\inf_{\theta\in\{0,1\}}J_{\gamma,\theta}^{\rm MOP}(\pi)$.
The infimum is taken after computing the source return in each world.
Histories include every chosen action and visited MDP state. The known
initial START can be omitted without losing information.

\begin{proposition}[MOP has permanent loss on a finite MDP]\label{prop:mop-finite-mdp}
For every $0<\gamma<1$, the worst-world objective and each per-world objective
have the same attained maximum over stationary policies. Every maximizer chooses READ at START
with probability $1/3$. For each such policy, deficiency to the native READ
experiment is exactly $1/3$ at every positive collection time and for the
entire infinite history. Thus the eventual deficiency remains $1/3$ along
any family of exact optimizers as $\gamma\uparrow1$. The pure READ policy is
sufficient and strictly finitarily dominates every MOP maximizer.
\end{proposition}
\begin{proof}
Let $p=\pi(\mathrm{READ}\mid\mathrm{START})$, and write
$h(p)=-p\log p-(1-p)\log(1-p)$, with $0\log0=0$.
Every reward after the first transition is zero: there is one available action
and its transition is deterministic. At START, READ has transition entropy
zero and NOISE has transition entropy $\log2$. Hence
\[
 J_{\gamma,\theta}^{\rm MOP}(\pi)=h(p)+(1-p)\log2
 =\log3-D\!\left((p,1-p)\,\middle\|\,(1/3,2/3)\right).
\]
Nonnegativity and the equality condition for relative entropy give the
unique maximizing startup distribution, attained by $p=1/3$. The return is
finite and independent of both the discount and the world, so taking the
infimum over worlds leaves the score and its maximizers unchanged.

READ reveals $\theta$ exactly. The NOISE branch is independent of $\theta$,
and subsequent states and actions add no information. A decoder copies the
revealed bit on the READ branch and guesses a fair bit on the NOISE branch;
its worst-world error is $(1-p)/2$. Conversely, the two history laws have total
variation distance $p$. Contractivity under every randomized decoder and the
triangle inequality force error at least $(1-p)/2$ when simulating the perfectly
distinguishing READ target. The argument applies to every positive finite history
and to measurable decoders of the full infinite history. Thus at every optimum
this deficiency is $1/3$. Pure READ reveals the world, from which a
world-independent decoder can simulate any finite native experiment, and is
therefore sufficient. The positive READ deficiency prevents reverse finitary
dominance, proving strict domination. The limit statement follows from the
constant deficiency.
\end{proof}

\section{State-visitation entropy in a three-state MDP}\label{app:state-entropy}

The MDP state space is $\{x_*,x_0,x_1\}$, with known initial state
$x_*$ (START), and the action set is $A=\{\mathrm{TEST},\mathrm{RANDOM}\}$.
There are two possible worlds, indexed by $\theta\in\{0,1\}$.
Their transition laws are
\[
\begin{aligned}
 P_\theta(x_\theta\mid x_*,\mathrm{TEST})&=1,\\
 P_\theta(x_i\mid x_*,\mathrm{RANDOM})&=\tfrac12\quad(i=0,1),\\
 P_\theta(x_i\mid x_i,a)&=1\quad(i=0,1;\ a\in A).
\end{aligned}
\]
The next MDP state is fully observed after each action. There is no reset
action. Policies may randomize and depend on the entire observed history;
all their actions remain in the collected history.

Fix $0<\gamma<1$. The objective is entropy of the normalized discounted state
occupancy, including the initial state:
\[
 d_{\pi,\gamma}^{\theta}(x)=(1-\gamma)\sum_{t\ge0}\gamma^t
          \Pr_\theta^\pi(S_t=x),\qquad
 J_\gamma^\theta(\pi)=-\sum_x d_{\pi,\gamma}^{\theta}(x)
                                  \log d_{\pi,\gamma}^{\theta}(x).
\]
We use natural logarithms and $0\log0=0$. Both the MDP state variable and the order of averaging and taking
entropy are retained. The objective is evaluated separately in each fixed MDP;
no prior mixture over possible worlds is taken inside the entropy. We select
policies by $J_\gamma(\pi)=\inf_{\theta\in\{0,1\}}J_\gamma^\theta(\pi)$.

Let $p=\Pr_\pi(A_1=\mathrm{TEST})$. After the first transition the MDP state
never changes, so the geometric series gives
\[
 \bigl(d_{\pi,\gamma}^{\theta}(x_*),
       d_{\pi,\gamma}^{\theta}(x_\theta),
       d_{\pi,\gamma}^{\theta}(x_{1-\theta})\bigr)
 =\left(1-\gamma,\frac{\gamma(1+p)}2,\frac{\gamma(1-p)}2\right).
\]
Writing $h(u)=-u\log u-(1-u)\log(1-u)$ for binary entropy, we obtain
\begin{equation}\label{eq:state-entropy-score}
 J_\gamma^\theta(\pi)=h(\gamma)+\gamma h\!\left(\frac{1+p}{2}\right).
\end{equation}
Since $h$ is uniquely maximized at $1/2$, the global maximizing policies are
exactly those with $p=0$. The maximum $h(\gamma)+\gamma\log2$ is attained.
The formula is independent of $\theta$, so the same characterization holds
for $J_\gamma$ and for each per-world objective, allowing arbitrary adaptive
continuation after the initial RANDOM action.

Let $T$ be the native one-transition experiment that chooses TEST. Its two
world rows have disjoint support. Under any maximizing policy, the first terminal
state is instead a fair draw independent of $\theta$. Every subsequent transition
is an absorbing self-loop, and the policy's action kernels are common to both
worlds. Induction therefore shows that the full history law is identical in the
two worlds at every finite collection time $t$. Any randomized decoder of that
history must produce a common distribution for both rows of $T$. The triangle
inequality gives error at least $1/2$; outputting a fair synthetic TEST answer
attains $1/2$. Consequently every maximizing policy satisfies
\[
 \delta(K_{\pi,t},T)=\tfrac12\quad(t\ge0),\qquad
 \inf_{t\ge0}\delta(K_{\pi,t},T)=\tfrac12.
\]
Choosing TEST reveals the world, from which a world-independent decoder can
simulate every finite native target. It is therefore sufficient and strictly
dominates every entropy optimizer. Because the optimizer's deficiency is exactly
$1/2$ for every proper discount, any family of exact optimizers retains that
error as $\gamma\uparrow1$; convergence to a limiting policy is unnecessary.
The statement concerns the policy evaluated on a single run. Training access to
repeated resets, used by the source's learning algorithm, is a different interface.

\section{DIAYN on actual observed states}\label{app:diayn}

Diversity Is All You Need (DIAYN) samples a skill label $Z$ from a fixed uniform
prior and uses a policy conditioned on that skill and the current MDP
state. Its information objective combines skill--state mutual information with
conditional action entropy \citep[Sections 3.1--3.2]{eysenbach2019diayn}.
We evaluate that objective under the limiting stationary state law from the
specified initial state. The source's Appendix B also uses stationary visitation
in its gridworld analysis. This example has no
reward deadline or discount.
The skill predictor, called the discriminator, takes the actual observed state
as input; neither actions nor the full interaction history are substituted for it.

\paragraph{Environment and objective.}
There are two possible worlds, indexed by $\theta\in\{0,1\}$,
and five fully observed states: the initial state $r$, two reading states
$R_0,R_1$, and two label states $L_0,L_1$. The three actions READ, LABEL$_0$,
and LABEL$_1$ remain available at every state. From $r$, READ moves to $R_\theta$
and LABEL$_j$ moves to $L_j$. Every other state is absorbing under every action.
Thus the unknown quantity is the transition law, while the current MDP
state is always observed.

A feasible skill family consists of all stationary randomized policies
$\pi_z(a\mid x)$, for $z\in\{0,1\}$. Draw $Z$ fairly once at startup, independently
of $\theta$, and execute the chosen policy. Write $B$ for its initial action and
$p_z(b)=\pi_z(b\mid r)$. For every time $t\ge1$, its state distribution is
\[
 d_z^\theta(R_\theta)=p_z(\mathrm{READ}),\qquad
 d_z^\theta(L_j)=p_z(\mathrm{LABEL}_j),
\]
with zero mass at the other states. This is both the limiting state law from
$r$ and an invariant distribution for the skill's transition kernel. It is a
distribution over randomized runs. The source's gridworld analysis additionally
assumes that state frequencies within a single episode equal the stationary
distribution. A single absorbing trajectory here need not have this whole
mixture as its empirical visitation frequencies.

Let $(Z,S,A)$ have joint mass
$\frac12 d_z^\theta(x)\pi_z(a\mid x)$. For an action-entropy coefficient
$\alpha>0$, define
\[
 F_\alpha^\theta(\pi)
   =I_\theta(Z;S)+\alpha H_\theta(A\mid S,Z).
\]
Entropies in this appendix use natural logarithms. The coefficient one is the
source's theoretical expression; its implementation uses a scaled entropy term,
including $\alpha=0.1$. Both quantities are evaluated in a fixed world
$\theta$, rather than after averaging the state distributions across worlds.
We select skill families by
$J_\alpha(\pi)=\inf_{\theta\in\{0,1\}}F_\alpha^\theta(\pi)$, keeping the fair
skill distribution fixed. The infimum is over worlds, not skills.

We evaluate the full action--state history of the policy that samples one skill
at startup and follows it. Its initial reading probability is
$s=\frac12[p_0(\mathrm{READ})+p_1(\mathrm{READ})]$.
Its target is the native one-transition READ experiment, which reveals $\theta$.

\begin{theorem}\label{thm:diayn-state-counterexample}
For every $\alpha>0$, the maximum of $J_\alpha$ over all feasible skill
families is attained and equals $\log 2+\alpha\log 3$. Each per-world objective
$F_\alpha^\theta$ has the same attained maximum. Under either criterion,
every maximizing family has $0\le s\le1/2$.
For the policy induced by each maximizing family, the deficiency to the READ
target at every positive collection time, and eventually, is $(1-s)/2$.
Every value in $[1/4,1/2]$ is attained as this deficiency by a maximizing family.
READ is sufficient and strictly finitarily dominates the policy induced by every
maximizing skill family.
\end{theorem}

\begin{proof}
There are two skills and three available actions, so
$I_\theta(Z;S)\le\log2$ and $H_\theta(A\mid S,Z)\le\log3$.
For skill $z$, take LABEL$_z$ at startup and thereafter choose actions uniformly.
The observed absorbing state identifies the skill, while the action entropy
remains $\log3$. Thus $J_\alpha$ and each $F_\alpha^\theta$ attain maximum
$M=\log2+\alpha\log3$. At any maximizer of $J_\alpha$,
$M=J_\alpha(\pi)\le F_\alpha^\theta(\pi)\le M$ for both worlds.
The witness's state and action distributions have these properties at every
positive time; the criterion is not restricted to a single scored decision.

At every maximum, the two component bounds must therefore be equalities.
In particular, $H_\theta(Z\mid S)=0$. For each fixed world, the initial branch
and its resulting state determine each other. Consequently, the root action
distributions $p_0$ and $p_1$ have disjoint supports. READ can have positive mass
under at most one skill, and its total probability is therefore at most $1/2$.

For any policy, READ reveals $\theta$ in the first state. Conditional on a
label branch, every subsequent state and adaptive action has a distribution
independent of $\theta$. Its full history is thus equivalent to revealing the
world with probability $s$ and otherwise reporting an erasure. This conclusion
also holds for arbitrary later continuations. A decoder copies a revealed
world and guesses fairly after an erasure, giving error $(1-s)/2$. Conversely,
the two source laws have total variation $s$, whereas the READ target laws have
total variation one. Data processing and the triangle inequality give
$1\le s+2\delta$, proving the matching lower bound at every positive time.
READ itself reveals the world and can simulate every finite native experiment,
so it is sufficient and strictly dominates any policy with $s<1$.

Finally, fix any $s\in[0,1/2]$. Let skill zero choose READ with probability
$2s$ and LABEL$_0$ otherwise; let skill one choose LABEL$_1$. Choose all later
actions uniformly. The root supports remain disjoint, both entropy bounds are
attained in both worlds, and the resulting policy reads with probability $s$.
This family also maximizes $J_\alpha$, supplying every claimed deficiency value
under both the worst-world and per-world criteria.
\end{proof}

\paragraph{Scope of the comparison.}
The skills accomplish their discrimination objective: the state identifies the
skill perfectly at every maximum. That property does not imply that their
collected evidence can reproduce the signal distributions of the native READ
experiment from the initial state.
The theorem concerns exact optimization of the source's information objective,
not convergence of its discriminator or soft actor-critic updates. The source's
learned discriminator $q(z\mid x)$ gives a variational lower bound, which equals
the information objective only when it realizes the skill posterior. In
particular, finite softmax logits need not attain the deterministic posteriors
in this example. We make no claim about attained neural-parameter optima or
training performance.

\section{Computing objective values of arbitrary policies}
\label{app:computability}
This example concerns computability of an objective on supplied policy
programs. A real value is \emph{computable uniformly in the policy} if one
algorithm takes a policy program and an integer $n$ and returns a rational
number within $2^{-n}$ of that value, whenever the program describes a total
computable policy. A policy is total computable when its next action can be
computed and returned at every finite history.

\begin{proposition}[Evaluation obstruction]\label{prop:policy-evaluation-obstruction}
There is a class of two deterministic computable worlds
with the following property. Let $W$ always wait and let $R$ reveal the unknown
bit at startup and then wait. No real-valued objective that is weakly monotone
for the finitary process preorder and satisfies $J(R)>J(W)$ is computable uniformly on
all total computable policies in this class.
\end{proposition}
\begin{proof}
The worlds are $\theta\in\{0,1\}$. WAIT returns zero and REVEAL returns
$\theta$, at every history. For a program $e$, define $\pi_e$ as follows.
At interaction $t$, simulate $e$ for its first $t$ computation steps. Reveal
if halting has been detected and the policy has not revealed before; otherwise
wait. This is a total computable action rule: every decision performs only a
bounded simulation, regardless of whether $e$ eventually halts.

If $e$ never halts, $\pi_e=W$. If it halts, revelation occurs at a finite
deterministic time. Both $R$ and $\pi_e$ can then reproduce every fixed finite-history
distribution of the other after sufficient collection: each can read the bit and
insert the required deterministic action and observation schedule.
Thus $\pi_e$ and $R$ are finitarily equivalent. Weak monotonicity forces
\[
 J(\pi_e)=\begin{cases}a=J(W),&e\text{ never halts},\\
 b=J(R),&e\text{ halts},\end{cases}\qquad a<b.
\]
If the proposed evaluator existed, evaluate the two fixed reference policies
to successively higher precision until their certified intervals are disjoint.
Because $a<b$, this terminates and supplies a positive rational separation.
Evaluating each $\pi_e$ more accurately than one quarter of that separation
would distinguish the two cases and decide halting. This contradicts the
classical undecidability result \citep[Section~8]{turing1936computable}.
\end{proof}

The objective need only be strict on the single comparison $R\succ_{\rm fin}W$.
The worlds themselves have trivial computable response functions, and each
constructed policy returns its action in finite time. The obstruction therefore
persists after restricting both worlds and policies to be computable.
It arises from unbounded postponement of acquisition, not difficulty in
predicting a finite interaction.

This result does not rule out an algorithm for selecting a good policy in a
restricted class. Immediate revelation is plainly sufficient in this example.
For a particular objective, finite-prefix approximations may still provide
computable increasing lower bounds. The obstruction concerns a uniform stopping
rule that approximates every total computable policy's objective value to any
requested accuracy.

\section{Finite planning comparison}
\label{app:empirical-overview}\label{app:finite-comparison}
The comparison in Section~\ref{sec:experiments} asks how well histories acquired
by optimizing different objectives reproduce alternative experiments when
collection is limited to a few interactions. We first select three-step policies
and evaluate their ability to simulate four-step experiments. We then select new
four- and five-step policies, compare them with fixed three-step reference
experiments, and measure the return cost of reproducing those references within
a fixed error tolerance. We optimize over full history-dependent policies using
the same response laws for each objective, with a uniform prior wherever one is
required. These are numerical planning comparisons, without sampled training or learned dynamics; neither
finite native objective implements the eventual score $J_w$.

\subsection{Worlds and available information}
Each of the 22 cases contains $q\in\{4,8\}$ possible worlds. One world $Q$,
fixed throughout a history, is represented by a hidden-state model with actions and
observations in $\{0,1\}$. Its state starts at zero. An interaction first draws
$s'$ from $T_Q(s'\mid s,a)$ and then observes $o$ from $Z_Q(o\mid s',a)$.
We compute policies using the supplied transition and observation arrays, with
uniform prior $\alpha(Q)=1/q$ for objectives that require one. During
interaction, each policy chooses actions from its action--observation history
and is given neither the physical state nor the actual world label. The policy
being evaluated and each target policy start from state zero in the same world;
target experiments do not continue from the state reached by the evaluated policy.

Seven classes have four worlds $(U,V)\in\{0,1\}^2$. Their names specify the
error probabilities $(e_U,e_V)$ of independent noisy bit observations:
\begin{itemize}
\item \textbf{Sensors} has one state. Actions 0 and 1 observe $U$ and $V$,
respectively. The error pairs are $(.1,.1),(.1,.3),(.25,.25)$.
\item \textbf{Delay} has states $0,1,2$. Action 0 advances to $\min(s+1,2)$;
its observation is a fair world-independent bit at state 1 and noisy $U$ at
state 2. Action 1 resets to state 0 and observes noisy $V$.
The error pairs are $(.1,.1),(.25,.1)$.
\item \textbf{Irreversible} has initial state 0 and absorbing states 1 and 2.
The first action selects state 1 or 2, where both actions subsequently observe
noisy $U$ or noisy $V$, respectively. The first observation already comes from
the selected state. The error pairs are $(.1,.1),(.1,.3)$.
\end{itemize}
The other fifteen classes have three hidden states. Each transition row is an
independent symmetric three-category Dirichlet draw with concentration $c$;
each observation row is an independent symmetric two-category draw with the
same $c$. This concentration is unrelated to the world prior. Three classes
have $q=4$ and $(\text{seed},c)=(91501,.2),(91502,1),(91503,5)$.
The other twelve use two draws for each
$(q,c)\in\{4,8\}\times\{.2,1,5\}$: seeds
2026096800--01, 2026096900--01, and 2026097000--01 for $q=4$;
2026100800--01, 2026100900--01, and 2026101000--01 for $q=8$.
NumPy's default random generator draws all transition rows before observation
rows. Saved arrays and hashes fix the instances independently of their seeds.
Figures use this case order. The cases are heterogeneous examples, and the
analyses were designed after inspecting them; they are not independent
replications or confirmation on unseen classes.

\subsection{Objectives, planning, and evaluation}
For a history $h$ with positive prior-predictive probability, the posterior is
\[
 p_h(Q)=\alpha(Q)K_{\pi,t}(Q,h)/\sum_{Q'}\alpha(Q')K_{\pi,t}(Q',h).
\]
The information and Brier returns are $I(Q;H_t)$ in bits and
\[
 R_B(\pi)=\mathbb E\!\left[\sum_Qp_{H_t}(Q)^2\right]-\sum_Q\alpha(Q)^2.
\]
These equal the expected sums of posterior relative-entropy increments and
squared posterior changes, respectively. Ordinary Brier is a quadratic proper
score \citep{brier1950verification,gneiting2007strictly}; here its predicted
variable is explicitly the whole world, not the next observation.

The categorical pseudo-count baseline instead scores raw observation symbols.
After $r$ observations its density is $\rho_r(o)=(N_r(o)+1)/(r+2)$, where
$N_r(o)$ counts occurrences of $o$. Updating once more on $o$ gives
$\rho'_r(o)=(N_r(o)+2)/(r+3)$. The recoding equations of
\citet[Sections 3.1 and 4]{bellemare2016unifying} therefore give pseudo-count
$\widehat N_r(o)=N_r(o)+1$ and pseudo-total $r+2$. The finite return is
$\mathbb E_{\alpha,\pi}\sum_{r=1}^t(1+N_{r-1}(O_r))^{-1/2}$, with scale one, no additional offset,
and counts reset at the start of each run. This categorical specialization does not test
the source's learned image-density implementation. A uniform control selects
each action with probability $1/2$ at every history.

To define the native objectives, let $\mathcal T_n$ be all deterministic
observation-contingent action trees of depth $n$. There are 128 for $n=3$.
For collection budget $t$, native mean and native maximum minimize, respectively,
\[
 L_{t,3}(\pi)=\frac1{128}\sum_{\sigma\in\mathcal T_3}
   \Def(K_{\pi,t},K_{\sigma,3}),\qquad
 M_{t,3}(\pi)=\max_{\sigma\in\mathcal T_3}
   \Def(K_{\pi,t},K_{\sigma,3}).
\]
Write $L_3=L_{3,3}$ and $M_3=M_{3,3}$. Each target gets its own randomized
decoder, which receives only the collected history and must work in every world.
Thus the mean averages over targets after taking each target's worst-world
error. These objectives include adaptive targets, not merely fixed action
sequences. Evaluation instead uses all 32,768 depth-four targets:
\begin{equation}\label{eq:finite-experiment-audit}
 A_{3,4}(\pi)=\max_{\sigma\in\mathcal T_4}
     \Def(K_{\pi,3},K_{\sigma,4}).
\end{equation}
Each evaluated policy still receives only three observations. A randomized target is a
world-independent mixture of deterministic trees; mixing their decoders preserves
an error bound. A shorter target follows by prefix projection. These reductions
cover randomized and shorter targets at the stated horizon, without extending
the audit to arbitrarily long experiments.

Information, Brier, and count policies are selected by backward dynamic
programming. Native planning uses linear programs in \emph{realization weights},
the products of the action probabilities assigned along each history.
Causal flow constraints ensure that these weights describe a policy. At each
reachable history, positive policy factors cancel from the posterior, so expected
posterior return is linear in the weights. Native maximum can therefore also be
minimized subject to a posterior return floor:
\[
 R(\pi)\ge R_{\max}-r(R_{\max}-R_{\min})-10^{-10},
 \qquad r\in\{.01,.05\},
\]
where both extrema range over the same three-step policy space. The 5\%-Brier control uses
constrained native optimization; it does not describe ordinary Brier optimization or
which nearly optimal policy a Brier learner would find.

The saved grid contains 427 selected-policy entries, identified by stable cell
identifiers in the supplementary data. Of the 339 ordinary entries, information,
Brier, count, and uniform each supply one policy per class (88 entries total).
Native mean supplies 116 and native maximum 135. Their multiple representatives
come from repeated planning on the same classes: random target orderings, or
orderings that place all fixed action sequences first, using seeds
923101--923103. These runs reach the same full 128-target objective.
Native maximum also includes 18 adaptive runs, which add targets with largest
current error and stop when the full-family upper bound and restricted-program
lower bound differ by at most $2\cdot10^{-7}$. The remaining 88 entries are
the four information/Brier allowance controls in each of 22 classes.
Thus the ordinary entries form $6\times22=132$ method/class groups, rather than
339 independent experimental cases. An \emph{envelope} encloses the saved
representatives in one group and their numerical bounds, not all optimizers or
sampling uncertainty. Its midpoint need not be attained, and separate audit and
return midpoints may correspond to different policies.

\subsection{Fixed-budget results and reward-matched controls}
\begin{figure}[p]
\centering
\includegraphics[width=\linewidth,height=.80\textheight,keepaspectratio]{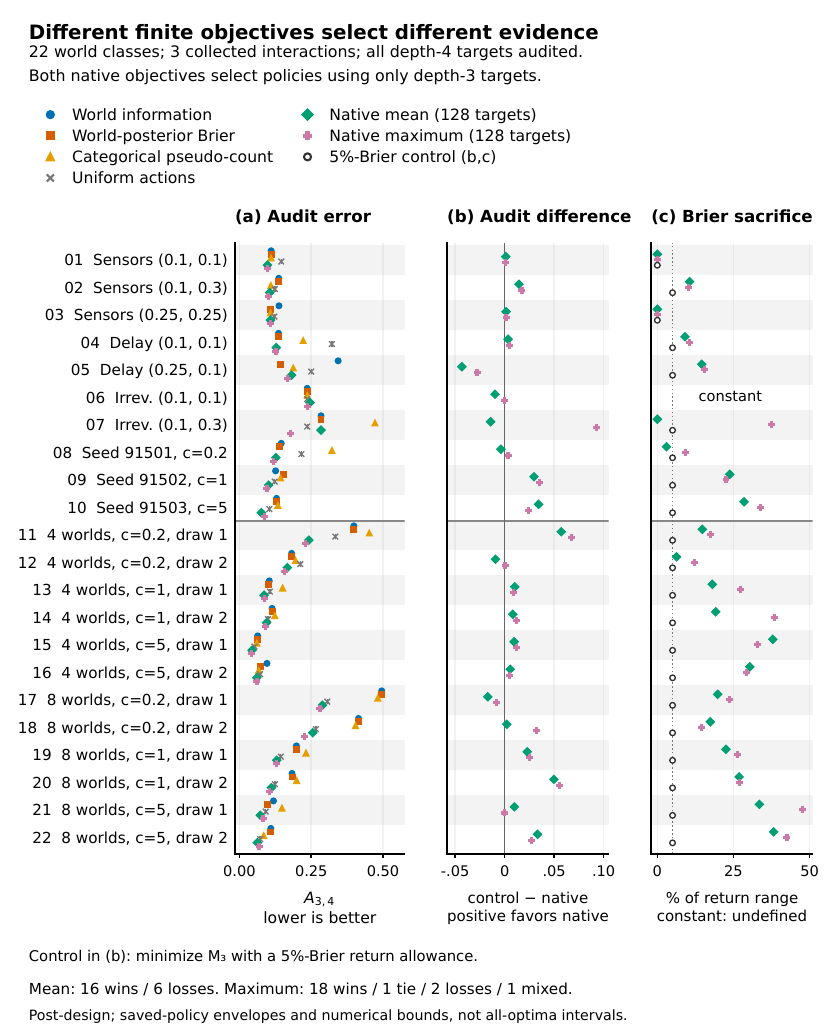}
\caption{\textbf{Fixed-budget simulation errors and returns.}
Three-step policies on all 22 classes: (a) hardest four-step target error,
(b) 5\%-Brier control minus native audit, and (c) Brier return forgone as a
fraction of its attainable range. The control minimizes the native maximum
subject to retaining Brier return within 5\% of its attainable range below the
maximum. Bars enclose saved representatives and numerical bounds, not all optima
or sampling uncertainty. Positive differences in (b) favor the native method.
The reward-matched comparison in Figure~\ref{fig:finite-objective-matched} instead
uses one fixed reference per class.}
\label{fig:finite-objective-comparison}
\end{figure}

Table~\ref{tab:finite-objective-results} and
Figure~\ref{fig:finite-objective-comparison}(a--c) report all 22 cases.
Native mean beats ordinary Brier on the audit in 18 cases, but often sacrifices
more Brier return than the 5\% control permits. The plotted sacrifice is
$(R_{\max}-R_B(\pi))/(R_{\max}-R_{\min})$; it is undefined when Brier return
is constant, as in Irreversible $(.1,.1)$. That case is omitted only from this
normalized fraction, not from simulation comparisons.

\begin{table}[!htbp]
\centering
\caption{\textbf{Worst four-step target error after three collected interactions.}
Counts favor the method named in the first column; every row retains all 22
classes. ``Mixed'' means signs differ across saved representatives. The last row
uses one fixed paired reference per class and cannot be read as changing only the
return floor in the preceding envelope comparison. The 5\%-Brier and matched
controls minimize the native depth-three maximum; they are not ordinary Brier optimizers.}
\label{tab:finite-objective-results}
\begin{tabular}{@{}llrrrr@{}}
\toprule
Method & Comparator & Wins & Ties & Losses & Mixed\\
\midrule
Native mean & Ordinary Brier & 18 & 2 & 2 & 0\\
Native maximum & Ordinary Brier & 19 & 1 & 2 & 0\\
Native mean & 5\%-Brier control & 16 & 0 & 6 & 0\\
Native maximum & 5\%-Brier control & 18 & 1 & 2 & 1\\
Fixed native mean & Its reward-matched control & 6 & 1 & 15 & 0\\
\bottomrule
\end{tabular}
\end{table}

If representatives give different comparison signs, the table reports ``mixed'';
numerical ties use tolerance $10^{-6}$.

\begin{figure}[p]
\centering
\includegraphics[width=\linewidth,height=.80\textheight,keepaspectratio]{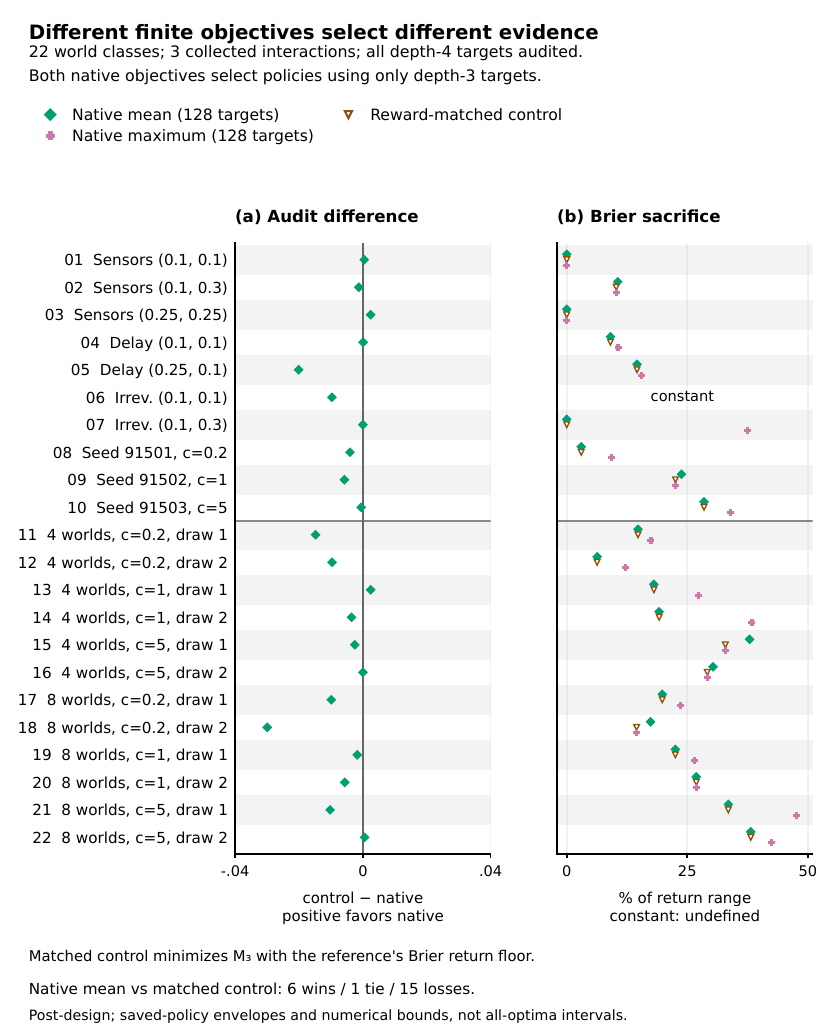}
\caption{\textbf{Retaining the reference's posterior return changes the comparison.}
(a) Each fixed native-mean reference versus native maximum constrained to retain
its Brier return, with allowance $10^{-10}$. Positive control-minus-reference
audit differences favor the reference: six wins, one numerical tie, and fifteen
losses. (b) Actual Brier sacrifice of the same paired policies, alongside the
original unconstrained native-maximum envelope. The return floor may be exceeded.
Both paired audits retain and replay every four-step target witness. These
post-design scalar results do not imply dominance: 21 pairs of experiments are
numerically incomparable and one is equivalent within tolerance.}
\label{fig:finite-objective-matched}
\end{figure}

The reward-matched diagnostic fixes the verified native-mean policy with the
lexically first saved cell ID in each class, without consulting the new audit values, and minimizes
$M_3$ subject to $R_B(\pi)\ge R_B(\pi_{\rm ref})-10^{-10}$.
The reference is feasible, so an exact control optimum cannot have worse $M_3$,
although this says nothing about $A_{3,4}$. The control may exceed its reward
floor. Native mean now has 6 wins, 1 numerical tie, and 15 losses, with median
control-minus-reference audit difference $-0.0030463$
(Figure~\ref{fig:finite-objective-matched}). Because this comparison fixes one
reference while the previous counts use envelopes, the changed counts do not
isolate the effect of changing the floor.

Directly comparing each selected pair in both directions separates these scalar
rankings from the experiment order. The original 88 pairs compare one selected
representative of each native method with ordinary Brier or the 5\%-Brier control;
84 are numerically incomparable and four equivalent within tolerance. Of the
22 matched pairs, 21 are incomparable and one equivalent. No strict Blackwell dominance
pair is reported. Information and Brier, in contrast, strictly respect Blackwell
improvement on these finite full-support world classes, so their exact optima
exclude strictly dominated feasible experiments. A maximum target error need only be
weakly monotone, and the finite mean does not inherit $J_w$'s general strictness
guarantee. The observed advantages therefore concern particular compromises,
without establishing a universally superior objective.

\subsection{More collection against fixed references}
\label{app:budget-comparison}
Within each class, fix the verified three-step policy with the lexically first
saved cell ID for each of the six methods. This gives six reference experiments
per class and $22\times6=132$ in total. Suppressing the class index, write $F_j$
for method $j$'s reference. The source $E_{m,t}$ is method $m$'s
selected experiment at budget $t$, and the measured error is
$d_{m,j}(t)=\Def(E_{m,t},F_j)$. At three steps we replay the fixed references;
at four and five steps we optimize new full-history policies using the same
response laws, prior, and initial states. These policies need not share prefixes, and
$d_{m,m}(3)=0$ is a self-comparison. Information, Brier, and count returns use the
new budget, while $L_{t,3}$ and $M_{t,3}$ retain their 128 targets and weights.
Every reference is covered by mixtures of those targets. For $t>3$, reversing
method names compares different sources and references, not reverse deficiencies
of one pair.

The $22\times6\times3=396$ source policies and their
$396\times6=2376$ within-class reference comparisons all received independently
checked numerical certificates. Certificate validity is separate from meeting a
simulation-error tolerance. A reference passes tolerance $\varepsilon$ when its upper error bound is
at most $\varepsilon-10^{-9}$, fails when its lower bound exceeds
$\varepsilon+10^{-9}$, and is otherwise unresolved. Each class passes only if
all six references pass, each with its own decoder. The thresholds $.01,.02,.05$
were fixed before these calculations. Table~\ref{tab:finite-budget-thresholds}
shows both the larger-budget gains and their dependence on the threshold: at
four steps Brier beats both native methods at $.02$, while at five steps the
native methods win more often. At $.01$, five-step Brier and native mean tie.
The supplementary data retain every case. A first successful
\emph{tested} budget is neither a minimum acquisition time nor a guarantee for
subsequently selected policies.

\begin{table}[!htbp]
\centering
\caption{\textbf{Classes reproducing all six fixed references.} Counts are verified
successes divided by checked classes, with all 22 classes checked for every method. Thresholds concern worst-world history
simulation error; they are not world-identification error rates.}
\label{tab:finite-budget-thresholds}
\begin{tabular}{@{}lrrr|rr@{}}
\toprule
& \multicolumn{3}{c}{Error tolerance $.02$} & \multicolumn{2}{c}{Five steps}\\
Objective & Three steps & Four steps & Five steps & $.01$ & $.05$\\
\midrule
Information & 1/22 & 5/22 & 8/22 & 6/22 & 13/22\\
Brier & 1/22 & 6/22 & 10/22 & 9/22 & 14/22\\
Categorical pseudo-count & 1/22 & 3/22 & 5/22 & 3/22 & 12/22\\
Uniform & 1/22 & 1/22 & 4/22 & 3/22 & 12/22\\
Native mean & 1/22 & 4/22 & 14/22 & 9/22 & 18/22\\
Native maximum & 1/22 & 4/22 & 16/22 & 10/22 & 19/22\\
\bottomrule
\end{tabular}
\end{table}

For example, in Delay $(.1,.1)$, Brier's errors to its own three-step reference
are $0$, $.072$, and numerically zero at budgets three, four, and five, even as
Brier return increases. The four-step policy selects a different tradeoff;
it does not continue an earlier policy and discard its history.

\subsection{Return cost of reproducing a reference}
To distinguish a necessary return sacrifice from selection among nearly optimal
policies, fix one reference $F$, budget $t$, and information or Brier return $R_t$.
If $R_t^*$ is the unconstrained maximum, define the \emph{compatibility cost}
\[
 C_{R,t}(F;\varepsilon)=R_t^*-
 \max_{\pi:\,\Def(K_{\pi,t},F)\le\varepsilon}R_t(\pi).
\]
We use $\varepsilon=.02$ and four pilot classes: Sensors $(.1,.1)$, Delay
$(.1,.1)$, Irreversible $(.1,.1)$, and the generated four-world class with seed
2026096800 and concentration $.2$. These were selected as the first entries of
their families before the new audits. There are $4\times2\times3\times6=144$
comparisons: four classes, two rewards, three budgets, and six references.
The prespecified eligibility rule selects a comparison when the ordinary
information- or Brier-selected policy at that budget has a checked lower
error bound above $.02+10^{-9}$ to the fixed reference. All 34 eligible comparisons
completed: six Sensors and six Irreversible comparisons have numerically zero
cost, while ten Delay and twelve generated-class comparisons have positive
lower bounds. Each reference may require a different compatible policy, and
these repeated comparisons within four classes are not a failure rate across
all 22. Favorable alternatives do not replace the primary policies.

The constrained problem is feasible by extending the three-step reference and
projecting its history. It is also a linear program: for a terminal history $h$
and decoder output $y$, combine realization weight $w_h$ and decoder probability
$G_h(y)$ as $z_{hy}=w_hG_h(y)$, with $z_{hy}\ge0$ and
$\sum_yz_{hy}=w_h$. If $p_Q(h)$ is the world's controlled-prefix mass, the
simulated target row is $\sum_hp_Q(h)z_{hy}$. Total-variation bounds in each
world, causal flow, and posterior return are then linear constraints or
objectives. Decoders receive the full observed history, with no hidden state,
world label, further interaction, or reset.

Policy and decoder replay lower-bounds the best compatible reward; a repaired
dual bound, adjusted to the intended cap $.02$, gives its upper bound.
An independent Bellman recurrence computes the unconstrained maximum.
Subtracting bounds gives the cost interval. Checks reconstruct rewards and
constraints from the saved response laws without invoking an optimizer.
A positive lower bound therefore gives evidence of required sacrifice under
the supplied numerical response laws, rather than only the loss of one chosen witness.

For Delay $(.1,.1)$ at four steps, recovering Brier's own reference lowers its
best gain from $.486322$ to approximately $.484039$, with cost bounded by
$[.0022829,.0022830]$, or $0.47\%$ of maximum gain.
This percentage uses maximum gain, unlike the attainable-range normalization
of the fixed-budget comparison.

In Irreversible $(.1,.1)$, a selected five-step information policy chooses one
bit and has error $.236$ to the balanced Brier
reference. A compatible witness mixes its first action with probabilities
approximately $(.517255,.482745)$, has decoder error below $10^{-12}$, and has
cost in $[0,5\cdot10^{-16}]$. The reference simulation error is sensitive to
numerical choices among nearly optimal policies. The four-step native-mean
representative similarly has mean target error $.236$ and Brier-reference
error $.236$, while a saved balanced native-maximum policy has the same mean
to within $6\cdot10^{-15}$ and Brier-reference error below $10^{-12}$.
These numerical agreements do not prove exact tied optima.

\begin{figure}[!htbp]
\centering
\includegraphics[width=\linewidth]{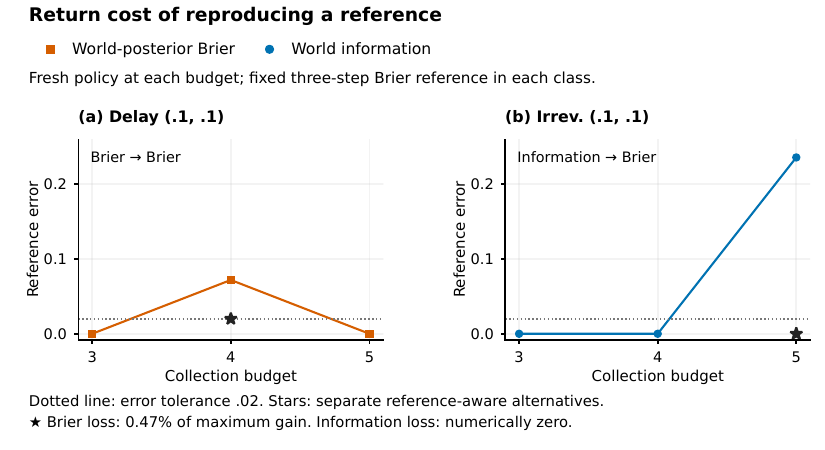}
\caption{\textbf{Return cost of reproducing a reference.}
(a) Brier to its Brier reference in Delay $(.1,.1)$.
(b) Information to the Brier reference in Irreversible $(.1,.1)$.
Stars show separate reference-aware alternatives: costs are $0.47\%$ of maximum
Brier gain and a numerically zero loss of information gain, respectively.
These policies are selected afresh at each budget; the lines do not track
one continuing policy.}
\label{fig:finite-objective-budget-summary}
\end{figure}

Finally, the balanced five-step native-maximum policy reproduces all six
references within $10^{-12}$, yet its hardest three-step target has error in
$[.2262799,.2262801]$. Success on the six references
therefore remains weaker than success on the entire native family.
Figure~\ref{fig:finite-objective-budget-summary} shows the two contrasting
compatibility examples.

\subsection{Verification, reproducibility, and scope}\label{app:finite-study-scope}
Independent checks reconstruct upper bounds from stochastic decoders and lower
bounds from bounded decision-loss witnesses, with accepted gaps at most
$2\cdot10^{-7}$. Source, policy, experiment-matrix, and certificate hashes
connect every summary to its numerical inputs. Full four-step witness replay
covers 63 audits: 18 posterior controls, one uniform audit, and both members of
all 22 matched pairs, each against all 32,768 targets.
The other ordinary profiles have checks of profile arithmetic and selected
planning, hardest-target, and full-revelation witnesses; they have not all
received exhaustive decoder replay. Neither exact expectations under the supplied laws nor
certificate checking removes floating-point arithmetic.

One Delay $(.25,.1)$ pair also has exact rational-table certificates: native-mean
cell 0014 and ordinary-Brier cell 0208. Interpret their saved binary64 entries
as exact rationals, normalize each row exactly, and remove globally zero columns.
For these probability tables $E_N,E_B$, the checked bounds are
\[
 .08782201\le\Def(E_N,E_B)\le .08782202,\qquad
 .02766393\le\Def(E_B,E_N)\le .02766394.
\]
The lower bounds hold against every real stochastic decoder, and explicit
decoders give the upper bounds. This proves the tables incomparable, without
certifying their correspondence to physical execution, policy optimality, or
the exhaustive four-step audit. A separate exact physical replay uses quantized
policy probabilities and is not substituted for these table bounds.

For the larger-budget study, each accepted source has replay of the state
transitions and observations, return checks, and decoder/decision-witness bounds.
Four uniform policies use explicit prefix decoders for their own references;
their other references use the same linear programs. All four still miss other
references at tolerance $.05$.
Native optimization at longer budgets uses a cutting-plane linear program with the same
128 targets and weights, retaining all target witnesses, dual constraints on
all syntactic histories, and final master certificates. Information and Brier
use full-tree dynamic programming. No target sampling is involved.

The supplementary artifact provides response arrays, selected policies, return
summaries, source hashes, numerical certificates, and reproduction instructions.
Complete tables retain all 2376 larger-budget reference audits and the four
uniform prefix-decoder checks.

Computation was not matched across methods. A separate computation-budget
study uses different targets and weights and is not pooled here.
An earlier ten-class benchmark optimized one fixed weighted loss
over sixteen native targets, with three collection steps and four worlds in
each class. Its evaluation averaged the optimal guessing accuracies for the
seven nonconstant binary labels on those worlds, modulo complements, under a
uniform prior. It then took the worst such average across the policies within
each objective's return allowance. On this metric, the native loss had zero
wins, five ties, and five losses against Brier at nominal zero regret, with
$10^{-8}$ numerical return slack and $10^{-6}$ comparison tolerance.
Its prespecified broad-benefit screen required a positive family-balanced
improvement over both information and Brier at zero and 1\% regret, with no
structural family losing more than one percentage point. It failed that screen.
The benchmark's protocol and all endpoints remain in the supplementary evidence.
Together, the studies illustrate tradeoffs among simulation errors and
intrinsic returns at finite budgets; they do not establish a generally superior
practical native objective, eventual dominance, or a characterization of all
reward optima.

\paragraph{Formal proofs.}\label{app:verification}
Lean proofs accompany the mathematical results. The companion artifact maps
claims to statements, assumptions, and declarations and gives build instructions.
Source fidelity is checked separately: proving a reward formula does not show
that it matches a published method. The compression constructions assume a
supplied complete optimal prefix interpreter; existence of that base interpreter
is not formalized here.

\end{document}